\documentclass[10pt,twocolumn,letterpaper]{article}

\usepackage{iccv}
\usepackage{times}
\usepackage{epsfig}
\usepackage{amsfonts}
\usepackage{graphicx}
\usepackage{amsmath}

\usepackage{amssymb}
\usepackage{booktabs}

\usepackage{dsfont}

\usepackage{cuted}
\usepackage{capt-of}

\usepackage{float}
\usepackage{subcaption}

\usepackage[font=small]{caption}

\usepackage[pagebackref=true,breaklinks=true,letterpaper=true,colorlinks,bookmarks=false]{hyperref}
\usepackage[capitalize]{cleveref}

\crefname{section}{Sec.}{Secs.}
\Crefname{section}{Section}{Sections}
\Crefname{table}{Table}{Tables}
\crefname{table}{Tab.}{Tabs.}

\makeatletter
\def\blfootnote{\gdef\@thefnmark{}\@footnotetext}
\makeatother

\newcommand{\VSPACEAMOUNT}{{\vspace{-1.5em}}}

\iccvfinalcopy 

\def\iccvPaperID{6657} 

\begin{document}

\title{Nerfbusters: Removing Ghostly Artifacts from Casually Captured NeRFs}

\author{Frederik Warburg$^{*2}$ \hspace{20mm} Ethan Weber$^{*2}$ \hspace{20mm} Matthew Tancik$^2$\\ \vspace{0mm} Aleksander Holynski$^{1,2}$ \hspace{15mm} Angjoo Kanazawa$^2$\vspace{5mm}\\$^1$Google Research\hspace{15mm}$^2$University of California, Berkeley}


\maketitle
\ificcvfinal\thispagestyle{empty}\fi

\begin{strip}\centering
\vspace*{-4em}  
\captionsetup{type=figure}
\includegraphics[width=\textwidth]{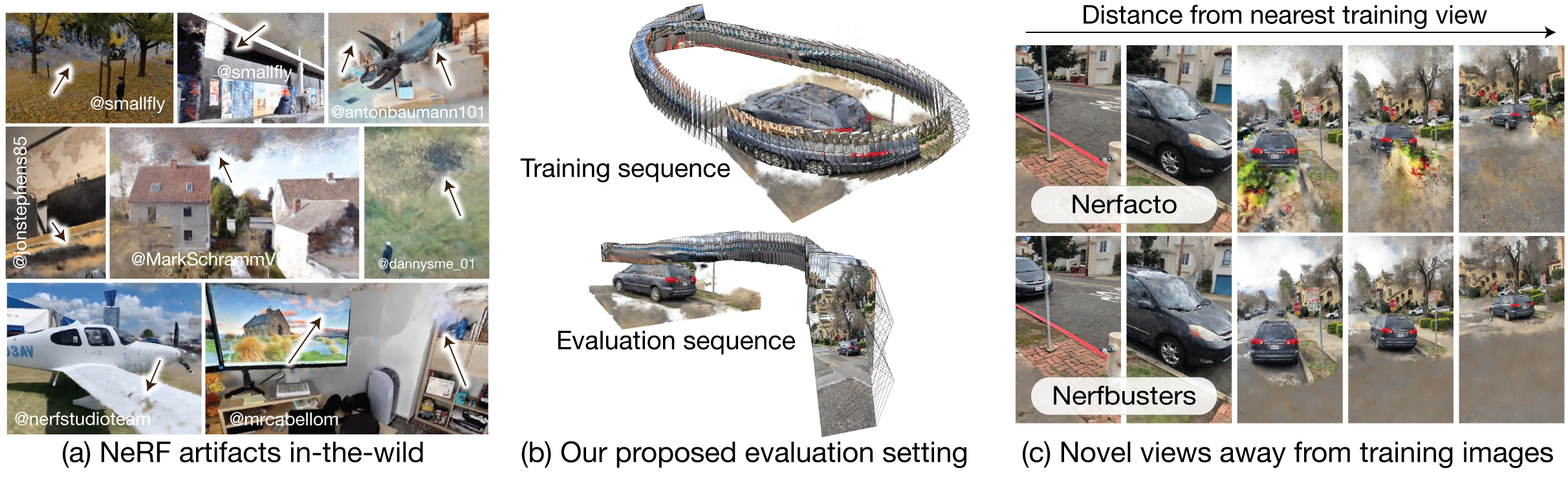}
\captionof{figure}{\textbf{Nerfbusters.} Rendering NeRFs at novel views far away from training views can result in artifacts, such as floaters or bad geometry. These artifacts are prevalent in in-the-wild captures (a) but are rarely seen in NeRF benchmarks, because evaluation views are often selected from the same camera path as the training views. We propose a new dataset of in-the-wild captures and a more realistic evaluation procedure (b), where each scene is captured by two paths: one for training and one for evaluation. We also propose Nerfbusters, a 3D diffusion-based method that improves scene geometry and reduces floaters (c), significantly improving upon existing regularizers in this more realistic evaluation setting. 
\label{fig:teaser}}
\end{strip}

\begin{abstract}
\vspace{-.5em}
Casually captured Neural Radiance Fields (NeRFs) suffer from artifacts such as floaters or flawed geometry when rendered outside the camera trajectory. Existing evaluation protocols often do not capture these effects, since they usually only assess image quality at every 8th frame of the training capture. 
To push forward progress in novel-view synthesis, we propose a new dataset and evaluation procedure, where two camera trajectories are recorded of the scene: one used for training, and the other for evaluation. In this more challenging in-the-wild setting, we find that existing hand-crafted regularizers do not remove floaters nor improve scene geometry. Thus, we propose a 3D diffusion-based method that leverages local 3D priors and a novel density-based score distillation sampling loss to discourage artifacts during NeRF optimization. We show that this data-driven prior removes floaters and improves scene geometry for casual captures.
\blfootnote{\hspace{-2em}*Denotes equal contribution}
\end{abstract}
\section{Introduction}

Casual captures of Neural Radiance Fields (NeRFs) \cite{mildenhall2021nerf} are usually of lower quality than most captures shown in NeRF papers. When a typical user (e.g., a hobbyist) captures a NeRFs, the ultimate objective is often to render a fly-through path from a considerably different set of viewpoints than the originally captured images. This large viewpoint change between training and rendering views usually reveals \emph{floater} artifacts and bad geometry, as shown in \cref{fig:teaser}a. 
One way to resolve these artifacts is to teach or otherwise encourage users to more extensively capture a scene, as is commonly done in apps such as Polycam\footnote{\url{https://poly.cam/}} and Luma\footnote{\url{https://lumalabs.ai/}}, which will direct users to make three circles at three different elevations looking inward at the object of interest. However, these capture processes can be tedious, and furthermore, users may not always follow complex capture instructions well enough to get an artifact-free capture. 

Another way to clean NeRF artifacts is to develop algorithms that allow for better out-of-distribution NeRF renderings. Prior work has explored ways of mitigating artifacts by using camera pose optimization~\cite{wang2021nerf,lin2021barf} to handle noisy camera poses, per-image appearance embeddings to handle changes in exposure~\cite{martin2021nerf}, or robust loss functions to handle transient occluders~\cite{sabour2023robustnerf}. However, while these techniques and others show improvements on standard benchmarks, most benchmarks focus on evaluating image quality at held-out frames from the training sequence, which is not usually representative of visual quality at novel viewpoints. 
\cref{fig:teaser}c shows how the Nerfacto method starts to degrade as the novel-view becomes more extreme.

In this paper, we propose both (1) a novel method for cleaning up casually captured NeRFs and (2) a new evaluation procedure for measuring the quality of a NeRF that better reflects rendered image quality at novel viewpoints. Our proposed evaluation protocol is to capture two videos: one for training a NeRF, and a second for novel-view evaluation (\cref{fig:teaser}b). Using the images from the second capture as ground-truth (as well as depth and normals extracted from a reconstruction on \emph{all} frames), we can compute a set of metrics on visible regions where we expect the scene to have been reasonably captured in the training sequence. Following this evaluation protocol, we capture a new dataset with $12$ scenes, each with two camera sequences for training and evaluation. 



We also propose \emph{Nerfbusters}, a method aimed at improving geometry for everyday NeRF captures by improving surface coherence, cleaning up floaters, and removing cloudy artifacts. Our method learns a local 3D geometric prior with a diffusion network trained on synthetic 3D data and uses this prior to encourage plausible geometry during NeRF optimization. Compared to global 3D priors, local geometry is simpler, category-agnostic, and more repeatable, making it suitable for arbitrary scenes and smaller-scale networks (a $28$ Mb U-Net effectively models the distribution of all plausible surface patches). Given this data-driven, local 3D prior, we 
use a novel unconditional Density Score Distillation Sampling (DSDS) loss to regularize the NeRF. We find that this technique removes floaters and makes the scene geometry crisper. To the best of our knowledge, we are the first to demonstrate that a learned local 3D prior can improve NeRFs. Empirically, we demonstrate that Nerfbusters achieves state-of-the-art performance for casual captures compared to other geometry regularizers.

We implement our evaluation procedure and Nerfbusters method in the open-source Nerfstudio repository~\cite{tancik2023nerfstudio}. The code and data can be found at \url{https://ethanweber.me/nerfbusters}.
\section{Related Work}


\begin{figure}
    \centering
    \includegraphics[width=\linewidth]{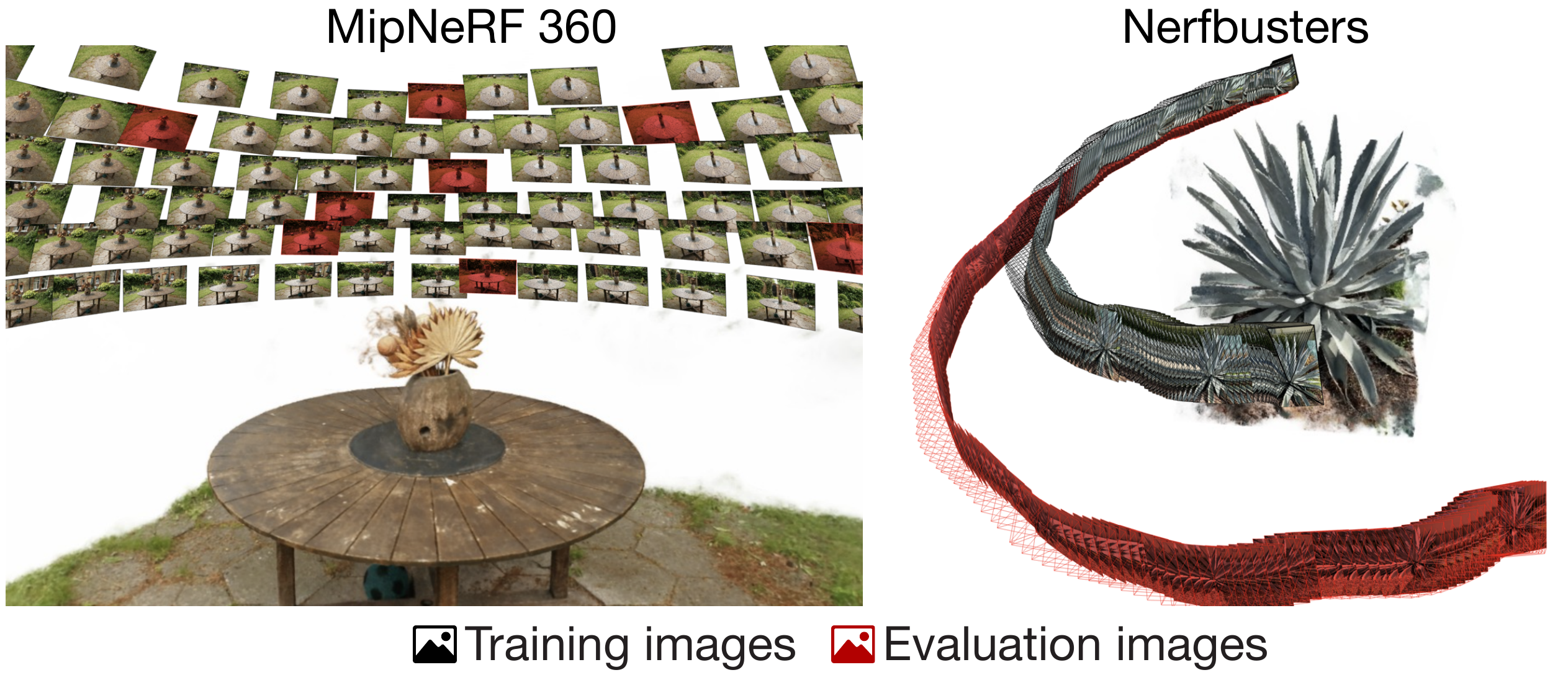}
    \caption{\textbf{Evaluation protocols.} Current evaluation of NeRFs (e.g., MipNeRF 360) measures render quality at every 8th frame of the captured (training) trajectory, thus only testing the model's ability to render views with small viewpoint changes. In contrast, we propose a new evaluation protocol, where two sequences are captured of the same scene: one for training and one for evaluation. Please see the supplementary material for plots showing the training and evaluation sequences for various NeRF datasets, including our proposed Nerfbusters Dataset.}
    \label{fig:related_work}
\end{figure}

\textbf{Evaluating NeRFs in-the-wild.} Early works in neural rendering \cite{mildenhall2019local}, including NeRF \cite{mildenhall2021nerf}, established an evaluation protocol for novel view synthesis, where every 8th frame from a camera trajectory is used for evaluation. Most follow-up works have adapted this protocol and demonstrated impressive results on forward-facing scenes in LLFF~\cite{mildenhall2019llff}, synthetic scenes \cite{mildenhall2021nerf}, or 360 scenes \cite{barron2022mip, reizenstein21co3d}. In these datasets, the training and evaluation views share camera trajectories, thus the methods are evaluated only for small viewpoint changes, as illustrated in \cref{fig:related_work}. In contrast, we propose to record two camera trajectories, one for training and one for evaluation. \cref{sec:appendix_datasets} compares existing datasets (synthetic scenes \cite{mildenhall2021nerf}, LLFF \cite{mildenhall2019llff}, MipNeRF 360 \cite{barron2022mip}, and Phototourism \cite{jin2021image}) with the proposed Nerfbusters dataset. We visualize the training and evaluation poses for each scene and quantify the difficulty of each dataset by computing the average rotation and translation difference between evaluation images and their closest training images. We find that viewpoint changes are very limited, and the proposed Nerfbusters dataset is much more challenging. Recently, Gao \etal \cite{gao2022monocular} revisited the evaluation process for dynamic NeRFs, also highlighting shortcomings in dynamic NeRF evaluation. NeRFs for extreme viewpoint changes and few-shot reconstruction have been explored on ShapeNet \cite{chang2015shapenet}, DTU~\cite{jensen2014large}, and CO3D~\cite{reizenstein21co3d}, where a few or just a single view is available during training. These works focus on the generalization and hallucination of unseen regions, and either assume a category-specific prior~\cite{zhou2022sparsefusion, yu2021pixelnerf} or focus on simple scenes~\cite{yu2021pixelnerf}. In contrast, our casual captures setting assumes that a $10-20$ second video is available at training time, better reflecting how people capture NeRFs. We then evaluate fly-throughs with extreme novel views on an entirely different video sequence, as illustrated in \cref{fig:related_work}.

\textbf{Diffusion models for 3D.} Recently, several works have proposed the use of diffusion models for 3D generation or manipulation~\cite{poole2022dreamfusion, wang2022score, muller2022diffrf, wynn-2023-diffusionerf}. These approaches can be divided into (1) methods that distill priors from existing 2D text-to-image diffusion models into a consistent 3D representation~\cite{poole2022dreamfusion, wang2022score}, and (2) methods that train a diffusion model to explicitly model 3D objects or scenes~\cite{muller2022diffrf}. These directions are complementary, where the former benefits from the sheer size of image datasets, and the latter from directly modeling 3D consistency. DreamFusion~\cite{poole2022dreamfusion} proposes Score Distillation Sampling (SDS), where the text-guided priors from a large text-to-image diffusion model can be used to estimate the gradient direction in optimization of a 3D scene. 
We take inspiration from the SDS optimization procedure but instead adapt it to supervise NeRF densities in an unconditional manner, directly on 3D density values. 
As the underlying model, we train a 3D diffusion model on local 3D cubes extracted from ShapeNet~\cite{chang2015shapenet} objects. 
We find the distribution of geometry (surfaces) within local cubes is significantly simpler than 2D natural images or global 3D objects, reducing the need for conditioning and high guidance weights. 
To the best of our knowledge, we are the first to suggest a learned 3D prior for category-agnostic, unbounded NeRFs.

\textbf{Data-driven local 3D priors.} Approaches for learning 3D geometry can be divided into local and global approaches, where global approaches reason about the entire scene, and local approaches decompose the scene into local surface patches. We learn a prior over local geometric structures, as local structures are simple, category-agnostic, and more repeatable than global structures \cite{chabra2020deep, mittal2022autosdf}. DeepLS \cite{chabra2020deep} proposes to decompose a DeepSDF \cite{park2019deepsdf} into local shapes, and finds that this simplifies the prior distribution that the network learns. Similarly, AutoSDF \cite{mittal2022autosdf} learns a local 3D prior and tries to learn the distribution over a 3D scene with an autoregressive transformer. We are inspired by their approach, but use a diffusion model rather than a VQ-VAE \cite{razavi2019generating, van2017neural}, and show that the learned prior can be used to regularize NeRF geometry.

\textbf{Regularizers in NeRFs.} Our work can be seen as a regularizer for NeRFs. Most existing regularizers are hand-crafted priors that encourage smoothness and discourage non-empty space. Plenoxels \cite{fridovich2022plenoxels} proposed a Total-Variation (TV) regularizer in 3D that penalizes the large changes between neighboring voxels. TV has also been applied in 2D rendered images in RegNeRF \cite{Niemeyer_2022_CVPR} and on the basis of factorized plenoptic fields \cite{chen2022tensorf, fridovich2023k}. Plenoctrees \cite{yu2021plenoctrees} proposed a sparsity loss that penalizes densities from randomly queried 3D locations in space. This sparsity loss removes densities unnecessary in explaining the training views. To avoid penalizing all densities equally, MIP-NeRF 360 \cite{barron2022mip} proposes a distortion loss
on accumulated weights that encourages surfaces to be sharp. Concurrent and most similar to our work, DiffusionRF \cite{wynn-2023-diffusionerf} proposes a data-driven RGB-D diffusion prior. The method trains a diffusion model on synthetic RGB-D examples and uses the learned prior to regularize a NeRF. In contrast to the proposed local 3D diffusion prior, operating in 2.5D comes with several disadvantages, namely 1) occlusions are not modeled, 2) the joint distribution of RGB-D images is more complex than that of 3D occupancy (and as a result requires more data for generalization), 3) and unlike a 3D diffusion model, it is not by definition 3D-consistent, \ie, the view consistency has to come from the NeRF rather than the regularizer. 

\section{Evaluation Procedure}\label{sec:evaluation_procedure}

\begin{figure}[t]
    \centering
    \includegraphics[width=\linewidth]{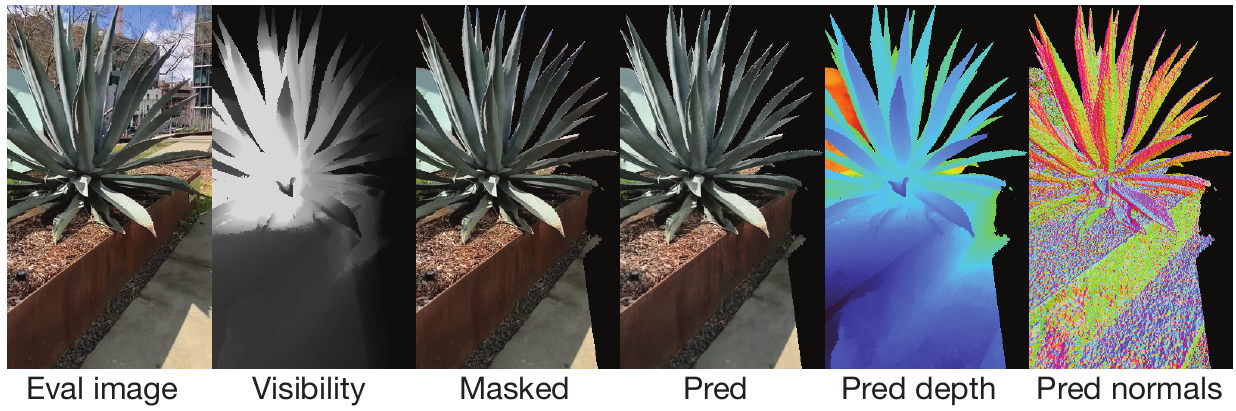}
    \caption{\textbf{Evaluation capture.} Here we show the data used in our evaluation protocol. The evaluation trajectory is a separate capture that is held out during optimization of the NeRF. Individual components shown here are further described in~\cref{sec:evaluation_procedure}.}
    \label{fig:pseudo_gt}
\VSPACEAMOUNT
\end{figure}

We propose an evaluation protocol that better reflects the rendering quality of a captured NeRF at novel viewpoints. Under this protocol, a scene should be captured by two videos, one for training and one for evaluation. Training videos should be around $10-20$ seconds, which is representative of what a user might do when prompted to scan an object or scene (as anything longer than this may reduce the appeal and practicality of NeRF captures). The second video should ideally capture a set of novel views that a user may wish to render. For everyday user captures, the second video is not needed, as it is only used as ground truth in evaluation. 
We record $12$ scenes (two videos each) following this protocol to construct our Nerfbusters Dataset. All videos were taken with a hand-held phone to best simulate a typical casual capture setup.

\textbf{Evaluating on casual captures.} The steps to create our evaluation data can be boiled down to the following straightforward steps:
\begin{enumerate}
\itemsep-0.4em
    \item Record a video to capture the scene (training split).
    \item Record a second video from a different set of viewpoints (evaluation split).
    \item Extract images from both videos and compute camera poses for all images.
    \item Train a ``pseudo ground truth'' model on a combination of both splits and save depth, normal, and visibility maps for evaluation viewpoints.
    \item Optimize a NeRF on the training split and evaluate at novel viewpoints (using the captured images and pseudo ground truth maps) from the evaluation split.
\end{enumerate}
 
In~\cref{fig:pseudo_gt}, we show an evaluation image and its visibility, depth, and normal maps. These pseudo ground truth properties are high-quality since they are extracted from a NeRF model that was trained on a combination of the training and evaluation views. The visibility map is computed by taking the depth map, back-projecting each pixel into a 3D point, and then counting how many training views observe that 3D point. Our final Nerfbusters dataset with associated visibility masks and processing code can be found at \url{https://ethanweber.me/nerfbusters}.

\textbf{Masking valid regions.} Rendering extreme novel views exposes parts of the scene that were not captured in the training views. As most existing NeRFs are not designed to hallucinate completely unseen views, we only evaluate regions in the evaluation images that were co-observed in the training views. We accomplish this by using visibility masks, which are defined to be regions that are either (1) not seen by any training views or (2) are predicted to be too far away (i.e., predicted depth $>$ distance threshold). We set this threshold to two times the largest distance between any two camera origins in both the training and evaluation splits. In the Nerfstudio codebase, this corresponds to a value of $2$ because camera poses are scaled to fit within a box with bounds (-1,-1,-1) and (1,1,1).

\textbf{Coverage.} We additionally report ``coverage'', which is the percent of evaluated pixels (i.e., after masking by both visibility~\cite{gao2022monocular} and depth) among all pixels in the evaluation viewpoints, a metric commonly reported in depth completion \cite{zhang2018deep, warburg2022self, warburg2022sparseformer}. For example, removing all densities and predicting infinite depth would result in zero coverage.

\textbf{Image quality and geometry metrics.} We use masked versions of PSNR, SSIM, and LPIPS for image quality. We also report on depth (MSE and mean abs. disparity difference) and normals (mean and median degrees, and the percent of valid pixels with normals $<$ 30 degrees). We report averages for all images in the Nerfbusters Dataset in Sec.~\ref{sec:experiments}.

\begin{figure}
    \centering
    \includegraphics[width=\linewidth]{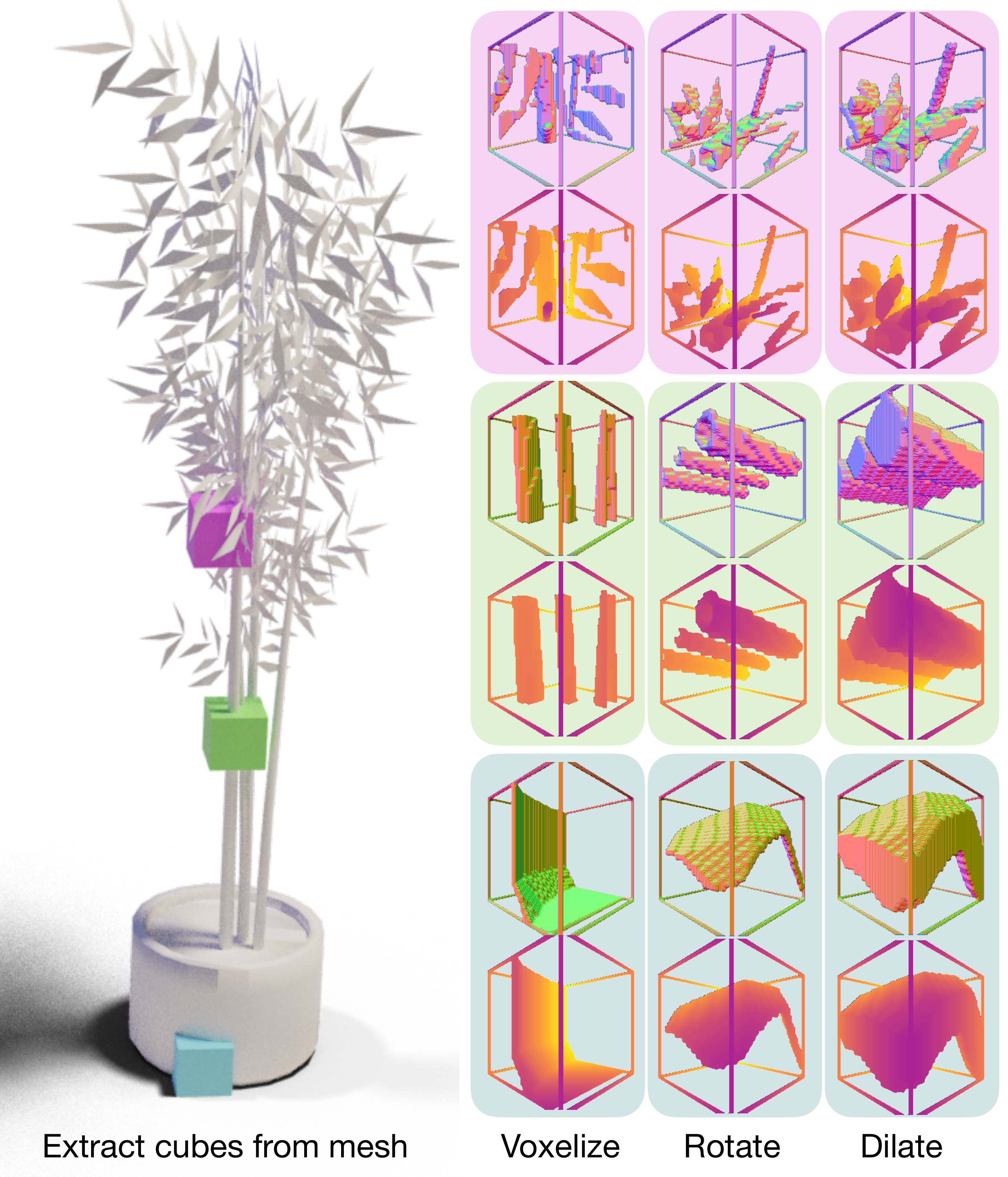}
    \caption{\textbf{Training data for Nerfbusters diffusion model.} Given a mesh, we extract local cubes scaled $1-10 \%$ of the mesh size. We voxelize these cubes with resolution $32^3$, and augment them with random rotations and random dilation. We illustrate each step with renderings of depth and normals from three cubes. The synthetic scenes from Shapenet offer a high variety in local cubes, containing both flat surfaces, round shapes, and fine structures. }
    \label{fig:data_processing}
\VSPACEAMOUNT
\end{figure}

\begin{figure*}[htp!]
    \centering
    \includegraphics[width=\textwidth]{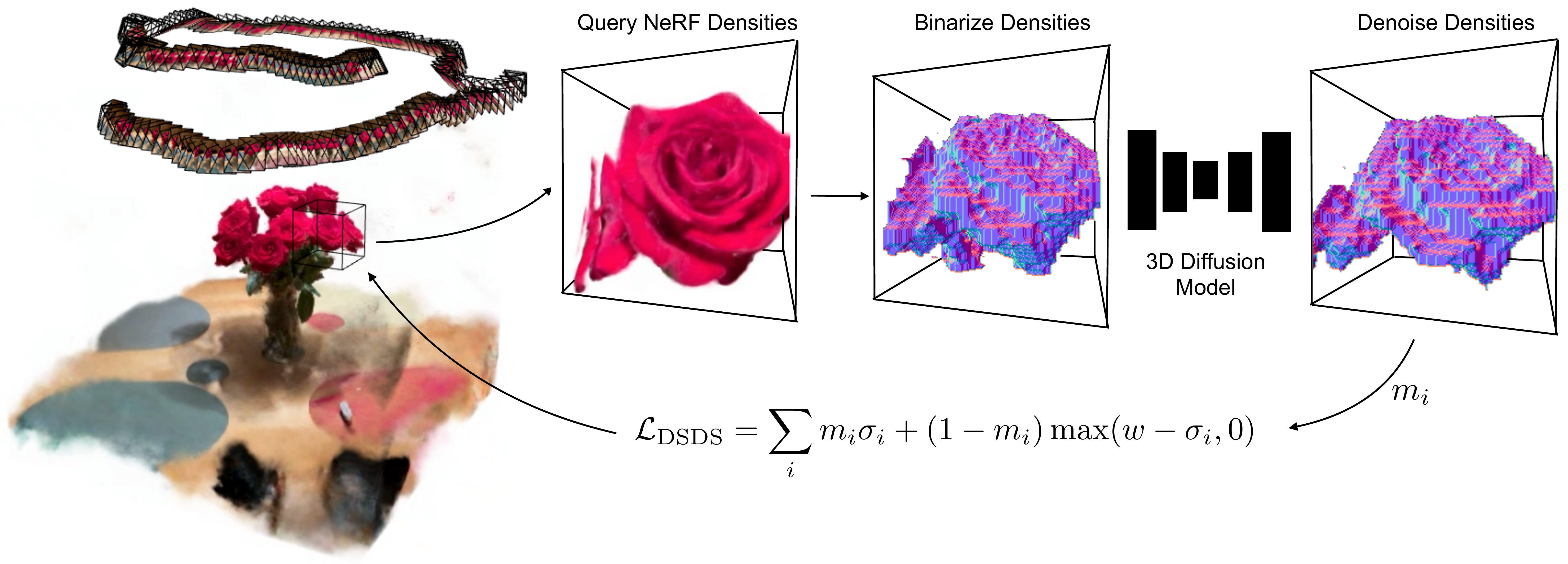}
    \caption{\textbf{Method overview.} We learn a local 3D prior with a diffusion model that regularizes the 3D geometry of NeRFs. We use importance sampling to query a $32^3$ cube of NeRF densities. We binarize these densities and perform one single denoising step using a pre-trained 3D diffusion model. With these denoised densities, we compute a Density Score Distillation Sampling (DSDS) loss that penalizes NeRF densities where the diffusion model predicts empty voxels and pushes the NeRF densities above the target $w$ where the diffusion model predicts occupied voxels $m = \mathds{1}\{x_0 < 0\}$.}
    \label{fig:model_overview}
\VSPACEAMOUNT
\end{figure*}

\section{Nerfbusters}

We propose a novel diffusion-based approach for regularizing scene geometry to improve NeRF rendering quality at novel viewpoints. Our method consists of two steps. First, we train a diffusion model to learn the distribution of 3D surface patches. This model is trained on synthetic data to unconditionally generate local 3D cubes. Second, we demonstrate an approach to apply this local prior in NeRF reconstruction of real 3D scenes. We do this by querying densities in local 3D patches in the scene during training and using a novel Density Score Distillation Score (DSDS) loss to regularize sampled density. This prior improves reconstructions in regions with sparse supervision signals and removes floaters. \cref{fig:model_overview} provides an overview of our pipeline.

\subsection{Data-driven 3D prior}

Following the recent process in generative models \cite{sohl2015deep, song2019generative, nichol2021improved, rombach2021highresolution, poole2022dreamfusion}, we formulate our local 3D prior as a denoising diffusion probabilistic model (DDPM) \cite{ho2020denoising}, which iteratively denoises a voxelized $32 \times 32 \times 32$ cube of occupancy $x$. Our diffusion model $\epsilon_{\theta}$ is trained with the loss:

\begin{equation}
    \mathcal{L}_\text{Diff} = \| \epsilon - \epsilon_\theta (\sqrt{\bar{\alpha}_t} x_0 + \sqrt{1 - \bar{\alpha}_t} \epsilon , t) \|^2_2
\end{equation}
where $t \sim \mathcal{U}(0, 1000)$, $\epsilon \sim \mathcal{N}(0, I)$ and $\bar{\alpha}$ follows a linear schedule that determines the amount of noise added at timestep $t$. We implement our diffusion model as a small 3D U-Net~\cite{ronneberger2015u} with only $7.2$M parameters ($28$MB).
We train the model on synthetic 3D cubes extracted from ShapeNet~\cite{chang2015shapenet} scenes.

\subsection{Curate synthetic 3D cubes}

We train our diffusion model on local cubes sampled from ShapeNet~\cite{chang2015shapenet}, illustrated in \cref{fig:data_processing}. To collect 3D cubes for training, we select a random ShapeNet mesh and extract $N$ local cube-bounded patches along the object surface with cube sizes varying between $1$-$10\%$ of the object's bounding volume. We voxelize these local samples at a resolution of $32^3$. We then augment the cubes with random amounts of rotation and dilation. This data processing pipeline is fast and performed online during training to increase the diversity of 3D cubes. We find that adjusting the thickness of the surface with dilation (rather than marking interior pixels as occupied) is faster and better defined for non-watertight meshes. \cref{fig:data_processing} illustrates the large diversity in the local cubes---some contain flat surfaces (bottom of the vase), round shapes (stem), and fine structures (leaves).

\subsection{Applying 3D prior in-the-wild} \label{sec:dsds}

We represent a 3D scene with a Neural Radiance Field (NeRF), \cite{mildenhall2021nerf} which takes a 3D point as input and outputs color and density, and is trained with differentiable volume rendering \cite{mildenhall2021nerf, max1995}. We build on the Nerfacto model from Nerfstudio \cite{tancik2023nerfstudio} that combines recent progress in NeRFs including hash grid encoding~\cite{muller2022instant}, proposal sampling~\cite{barron2021mip}, per-image-appearance optimization~\cite{martin2021nerf}, and scene contraction~\cite{barron2021mip}. Although Nerfacto has been optimized for in-the-wild image captures, it still reveals floaters when rendered from novel views. To address these issues, 
we propose a novel regularization strategy that includes (1) a mechanism for sampling cubes of occupancy from non-empty parts of the NeRF and (2) a novel Density Score Distillation Sampling (DSDS) loss that encourages sampled occupancy to agree with the learned geometry prior of the diffusion model. 

\textbf{Cube importance sampling. } Since the NeRF represents a density field, we can query cubes of occupancy in 3D space at any size, location, and resolution. 
Sampling these cubes uniformly from the NeRF volume is inefficient, particularly if much of the scene is empty. To enable more efficient sampling, we store a low-resolution grid of either \textit{accumulation weights} or \textit{densities}, which can be used to inform the probability with which to sample different positions in the scene. This grid is jointly updated with exponential moving average (EMA) decay over the course of NeRF training, such that regions with deleted floaters are not repeatedly sampled during later stages of training.  
The choice between the use of accumulation weights and densities has associated trade-offs: using \textit{accumulation weights} yields cubes sampled mostly on frequently seen surfaces, whereas using \textit{densities} enables sampling of occluded regions. In practice, our experiments use a grid of densities, clamped to $[0,1]$ to avoid a few densities dominating the sampling probability. This importance sampling method comes with almost no added cost since we store the densities or weights along the rays already used for volume rendering, and use a small $20^3$ grid. This grid informs the selection of the 3D cube center, and the size is randomly chosen in a range of $1$-$10\%$ of the scene, with a voxel resolution of $32^3$.

\textbf{Density Score Distillation Sampling (DSDS).} Our diffusion model is trained on discretized synthetic data in $\{-1, 1\}$ indicating free or occupied space, respectively. NeRF densities, on the other hand, are in $[0, \infty)$, where low densities indicate free space and larger densities mean more occupied space. In practice, we observe that densities less than $0.01$ are mostly free space, whereas occupied space have density values ranging from $[0.01, 2000]$. We propose a Density Score Distillation Sampling (DSDS) loss that handles the domain gap between the densities.

Given a cube of NeRF densities $\sigma$, we discretize the densities into binary occupancy: $x_t = 1$ if $\sigma > \tau$ else $-1$ for diffusion timestep $t$, where $\tau$ is a hyperparameter that decides at what density to consider a voxel for empty or occupied. The Nerfbusters diffusion model then predicts the denoised cube $x_0$. The timestep $t$ is a hyperparameter that determines how much noise the diffusion model should remove and can be interpreted as a learning rate. In practice, we choose a small $t \in [10, 50]$. With the denoised cube $x_0$, we penalize NeRF densities that the diffusion model predicts as empty or increase densities that the diffusion model predicts as occupied with:

\begin{equation}
    \mathcal{L}_{\text{DSDS}} = \sum_i m_i \sigma_i + (1 - m_i) \max(w - \sigma_i, 0),
\end{equation}

where $m=\mathds{1}\{x_0 < 0\}$ is a mask based on the denoised predictions. We penalize densities where the diffusion model predicts emptiness and increase densities where the model predicts occupancy. $w$ is a hyperparameter that determines how much to increase the densities in occupied space. The $\max$ operator ensures that no loss is applied if an occupied voxel already has a density above $w$. Similar to SDS \cite{poole2022dreamfusion, wang2022score}, the DSDS loss distills the diffusion prior with a single forward pass and without backpropagating through the diffusion model. Unlike SDS, our DSDS loss does not add noise to the original sampled occupancy before providing it to the diffusion model. While one may imagine that occupancy grids sampled during the NeRF optimization process do not exactly match the distribution of noised cubes used in training, we find that the denoising process nevertheless produces plausible samples that are both on the manifold of clean surfaces and similar in content to the input cubes.

\textit{Why not just...} use a differentiable function to convert densities to the valid range of the diffusion model, then compute the SDS loss \cite{poole2022dreamfusion, wang2022score}, and then backpropagate through the activation function? This would require a function $s: \sigma \rightarrow x_t$ to map $s(0) = -1$, $s(\tau) = 0$, and $s(2 \tau) = 1$, where $\tau$ is the crossing value where densities begin to be occupied. A scaled and shifted sigmoid function or a clamped linear function satisfies these requirements, but both have very steep gradients in some regions and no gradients in other regions, resulting in issues when backpropagating. In contrast, DSDS has gradients for any density predicted to be empty or occupied. In practice, we set $\tau = w = 0.01$ meaning our method deletes densities at points predicted to be empty and otherwise leaves the points unconstrained for the NeRF RGB loss to freely optimize.

\textit{Why not just...} use accumulated weights, which are in the range [0, 1]?
Weights are more well-behaved than densities but more expensive to compute as they require shooting a ray through the scene, evaluating and accumulating the densities along a ray. This results in significantly more function calls, but more fundamentally, requires one to specify a view from which to shoot the rays. This limits the diffusion prior to improving regions that are visible regions from the chosen view. A similar issue arises when using 2D or 2.5D priors \cite{Niemeyer_2022_CVPR, wynn-2023-diffusionerf}, where they may not regularize occluded regions unless viewpoints are chosen in a scene-specific way.

\begin{figure}[t]
    \centering
    \includegraphics[width=\linewidth]{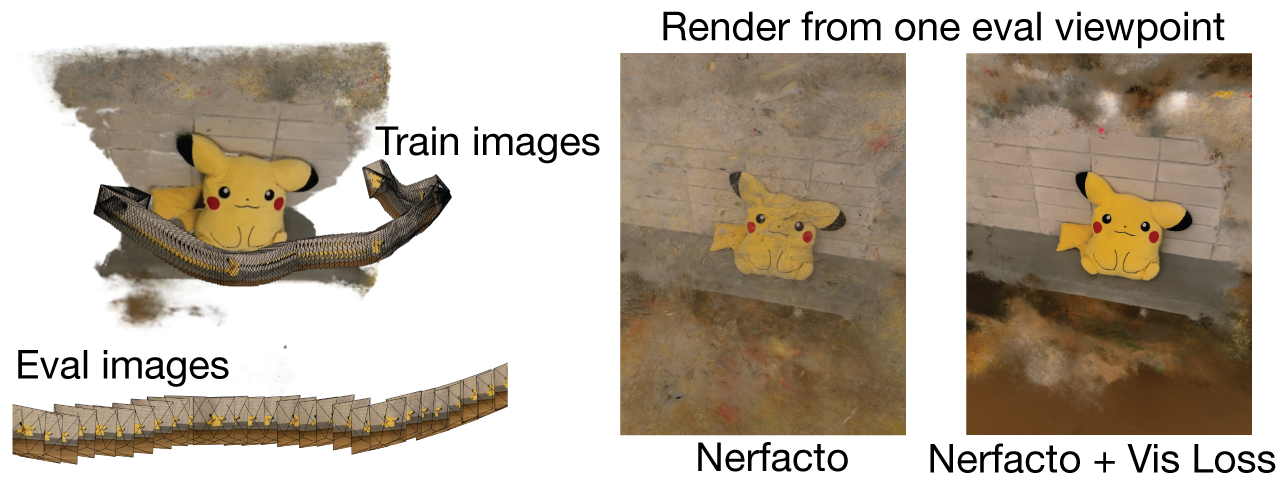}
    \caption{\textbf{Visibility loss.} Our visibility loss enables stepping behind or outside the training camera frustums. We accomplish this by supervising densities to be low when not seen by at least one training view. Other solutions would be to store an occupancy grid~\cite{muller2022instant} or compute ray-frustum intersection tests during rendering. Our solution is easy to implement and applicable to any NeRF.}
    \label{fig:visibility_loss}
\VSPACEAMOUNT
\end{figure}

\subsection{Visibility Loss}

Our proposed local 3D diffusion model improves scene geometry and removes floaters, but it requires decent starting densities since it operates locally and thus needs contextual information to ground its denoising steps. To this end, we propose a simple loss that penalizes densities at 3D locations that are not seen by multiple training views. We find this simple regularizer effective in removing floaters from regions outside the convex hull of the training views. We define our visibility loss as

\begin{equation}
    \mathcal{L}_{\text{vis}} = \sum_i V(q_i) f_{\sigma}( q_i)
\end{equation}

where $f_{\sigma}(q_i) = \sigma_i$ is the NeRF density at the 3D location $q_i$, and $V(q_i) = \mathds{1}\{ \sum_{j=1} v_{ij} < 1\}$ indicates if the location is not visible from any training views. We approximate the visibility $v_{ij} \in \{0, 1\}$ of the $i$'th 3D location in the $j$'th training view with a frustum check. This approximation does not handle occlusions, instead overestimates the number of views a location is visible from. This loss penalizes densities in regions not seen by training images.

In practice, we implement this by defining a single tight sphere around our training images and render batches of rays that shoot from a random location on the sphere surface, through the center of the scene, and far off into the distance. We render rays with Nerfacto and apply this loss to the sampled points. Nerfacto uses a proposal sampler~\cite{barron2022mip} to importance sample around surfaces, so our loss is effective in quickly culling away any floating artifacts with high density outside visible regions. See \cref{fig:visibility_loss} for a qualitative result where we render from behind training images.

\begin{figure*}[htp!]
    \centering
    \includegraphics[width=0.85\textwidth]{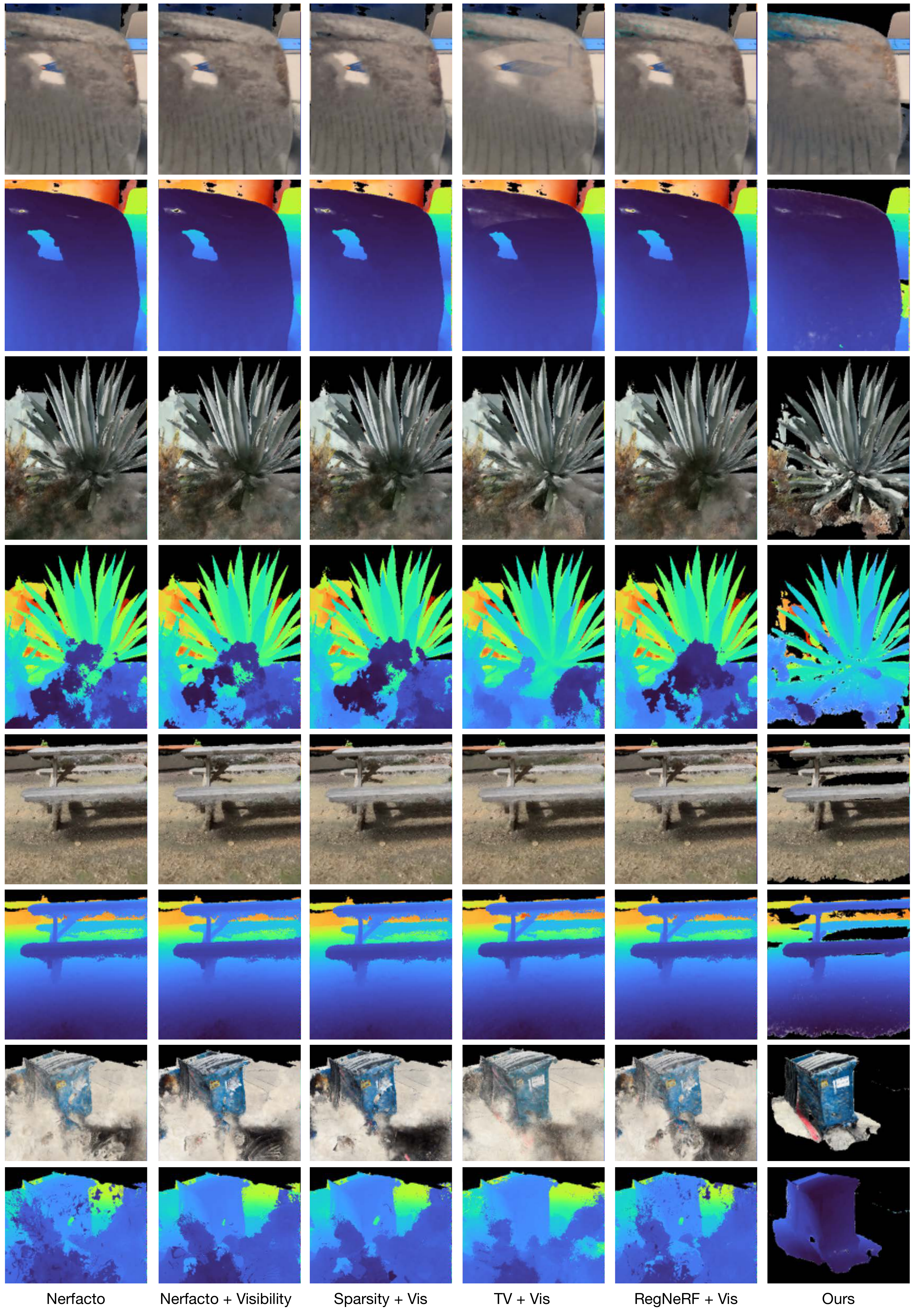}
    \caption{\textbf{Qualitative results.} NeRFs suffer from floaters and bad geometry when rendered away from training views. We show rendered RGB and depth from four scenes. In contrast to existing hand-crafted regularizers, our proposed diffusion prior fills holes (first scene), removes floaters (second and fourth scenes), and improves geometry (all scenes). Please see the associated website for video results on our evaluation splits.}
    \label{fig:qualitative}
\end{figure*}
\begin{table*}[]
\small
    \centering
    \begin{tabular}{l|lll|ll|lllll|l} 
        \toprule
        
 & PSNR $\uparrow$ & SSIM $\uparrow$ & LPIPS $\downarrow$ & Depth $\downarrow$ & Disp. $\downarrow$ & Mean $^{\circ}$ $\downarrow$ & Median $^{\circ}$ $\downarrow$ & \% $30^{\circ}$ $\uparrow$ & Coverage $\uparrow$ \\
 \midrule
        
 Nerfacto Pseudo GT & 25.98 & 0.8591 & 0.1019 & 0.0 & 0.0 & 0.0 & 0.0 & 1.0 & 0.893 \\ 
 Nerfacto           & 17.00 & 0.5267 & 0.3800 & 126.277 & 1.510 & 60.63 & 54.638 & 0.254 & \textbf{0.896} \\ 
  + Visibility Loss & 17.81 & 0.5538 & 0.3432 & 100.057 & 1.041 & 57.73 & 51.335 & 0.280 & 0.854 \\ 
  + Vis + Sparsity~\cite{yu2021plenoctrees}  & 17.81 & 0.5536 & 0.3445 & 92.168 & 1.145 & 57.77 & 51.399 & 0.280 & 0.854 \\ 
  + Vis + TV~\cite{fridovich2022plenoxels}        & 17.84 & 0.5617 & 0.3409 & 74.015 & 0.382 & 61.93 & 56.164 & 0.242 & 0.843 \\ 
  + Vis + RegNeRF~\cite{Niemeyer_2022_CVPR}   & 17.49 & 0.5396 & 0.3585 & 182.447 & 1.200 & 59.39 & 53.267 & 0.268 & 0.858 \\ 
  + Vis + DSDS (Ours)      & \textbf{17.99} & \textbf{0.6060} & \textbf{0.2496} & \textbf{54.453} & \textbf{0.114} & \textbf{54.77} & \textbf{47.981} & \textbf{0.295} & 0.630 \\ 

        \bottomrule
    \end{tabular}
    \caption{\textbf{Quantitative evaluation.} NeRFs suffer when rendered away from the training trajectories. Existing regularizers do not suffice to improve the geometry. Nerfbusters learns a local 3D prior with a diffusion model, which removes floaters and improves the scene geometry. Results are averaged across $12$ scenes.}
    \label{tab:results}
\VSPACEAMOUNT
\end{table*}

\begin{table}[]
\small
\centering
\resizebox{\linewidth}{!}{\begin{tabular}{l|ll|l|l|l}
\multicolumn{6}{c}{\textbf{Cube sampling strategies}} \\
\toprule
 & PSNR & SSIM & Disp. & Mean $^{\circ}$ & Cov. \\ \midrule
Uniform               & 14.61 & 0.4276 &    10.288    & 61.52 & \textbf{0.886} \\ 
Densities $\sigma$   & \textbf{16.46} & \textbf{0.5086} &    \textbf{0.081}     & \textbf{49.21} & 0.606 \\ 
Weights              & 15.86 & 0.4466 &    0.112       & 53.09 & 0.634 \\ 
\midrule
\multicolumn{6}{c}{\textbf{Activation functions}} \\
\toprule
 & PSNR & SSIM & Disp. & Mean $^{\circ}$ & Cov. \\ \midrule
 Clamp+SDS             & 12.53 & 0.2652 &      2.065     & 87.33 & \textbf{1.000} \\ 
 Sigmoid+SDS           & 12.53 & 0.2652 &       2.065          & 87.33 & \textbf{1.000} \\ 
 $\sigma_{\tau}$+DSDS        & \textbf{15.86} & \textbf{0.4466} &      \textbf{0.112}      & \textbf{53.09} & 0.634 \\ 
\midrule
\multicolumn{6}{c}{\textbf{Cube size range as \% of scene}} \\
\toprule
 & PSNR & SSIM & Disp. & Mean $^{\circ}$ & Cov. \\ \midrule
1-20\%          & \textbf{17.05} & \textbf{0.5005} &    \textbf{0.083}      & 54.87 & 0.600 \\ 
10-20\%          & 16.93 & 0.4884 &    0.090       & \textbf{50.78} & \textbf{0.640} \\ 
1-10\%             & 15.86 & 0.4466 &     0.112     & 53.09 & 0.634 \\ 
\bottomrule
\end{tabular}}
\caption{\textbf{Ablation study.} Ablation on the ``garbage'' scene for different settings of using our 3D prior as a NeRF loss. Cube sampling refers to uniformly sampling the entire scene versus importance sampling with accumulated weights or densities.}
\label{tab:nerf_ablation_study}
\VSPACEAMOUNT
\end{table}

\section{Experiments in-the-wild}\label{sec:experiments}

In this section, we follow our proposed evaluation protocol described in~Sec.~\ref{sec:evaluation_procedure} to ablate and compare the proposed method with a number of common regularizers aimed at cleaning up NeRFs.

\begin{figure}[t]
    \centering
    \includegraphics[width=1.0\linewidth]{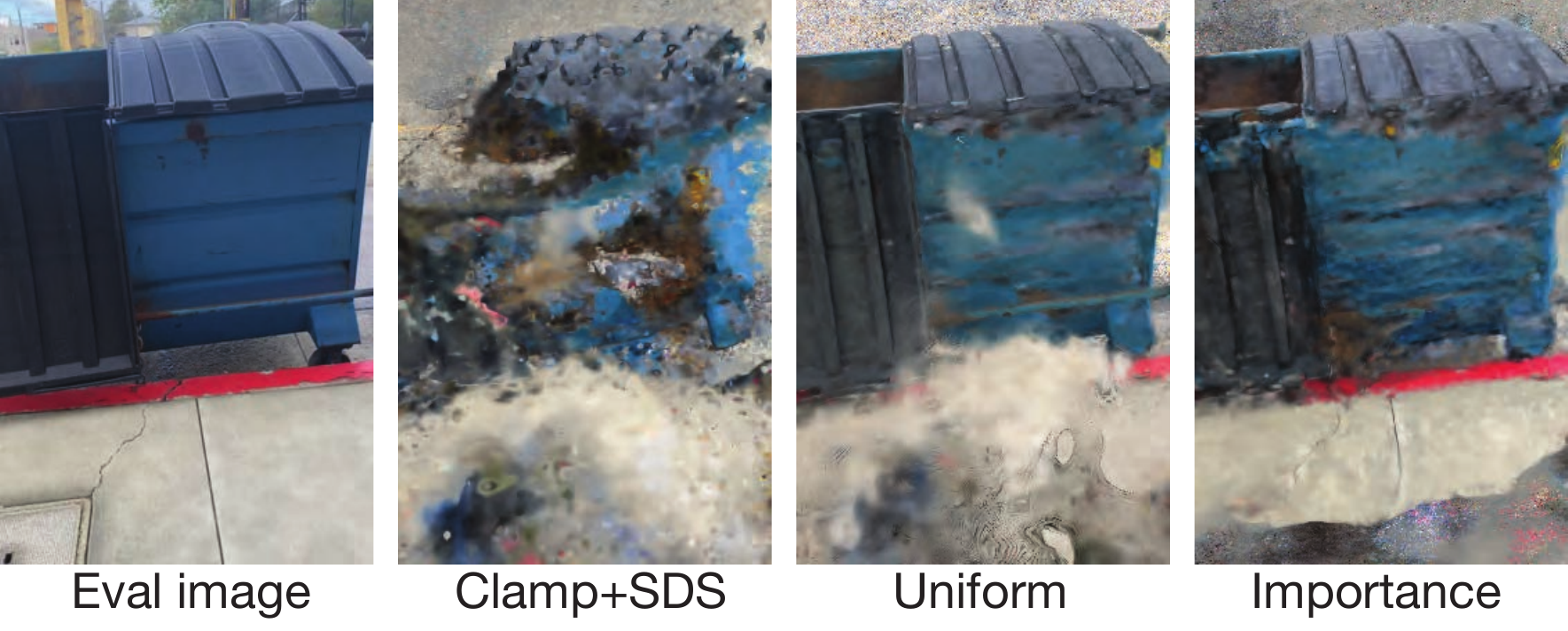}
    \caption{\textbf{Ablations results.} Using a simple activation function and SDS results in a not-well-behaved gradient signal, increasing the number of floaters in the scene. Importance sampling more effectively applies the 3D cube loss in space, cleaning up floaters and improving the scene geometry. }
    \label{fig:ablations}
\VSPACEAMOUNT
\end{figure}

\textbf{Implementation details.} For each experiment, we use the Nerfacto model within the Nerfstudio~\cite{tancik2023nerfstudio} codebase. We turn off pose estimation for evaluation purposes and then train Nerfacto for $30$K iterations which takes up to half an hour. We then fine-tune from this checkpoint with different regularizer methods. We compare the proposed method with vanilla Nerfacto, Nerfacto with the proposed visibility loss, Nerfacto with our visibility loss and 3D sparsity loss \cite{yu2021plenoctrees}, 3D TV regularization \cite{fridovich2022plenoxels}, and 2D TV (as in RegNeRF~\cite{Niemeyer_2022_CVPR}). Our implementations also use the distortion loss~\cite{barron2022mip} which is on by default with Nerfacto. All methods are effective within the first $1$K iterations of fine-tuning ($\sim$4 minutes on an NVIDIA RTX A5000 for Nerfbusters), but we train for $5$K iterations. For the 3D baselines, we sample 40 $32^3$ cubes per iteration and for the 2D baseline RegNeRF, we render ten $32^2$ patches. The usual NeRF reconstruction loss is also applied during fine-tuning with $4096$ rays per batch.

\textbf{Results.} \cref{tab:results} shows that visibility loss improves vanilla Nerfacto across all quality metrics. Existing hand-crafted regularizers do not improve upon this baseline. In contrast, our data-driven local diffusion prior removes floaters and improves the scene geometry, yielding state-of-the-art results on these challenging casual captures. The proposed method deletes floaters, and thus we find that it has lower coverage than the baselines. \cref{sec:appendix_results} shows per scene results. \cref{fig:qualitative} shows a qualitative comparison of the methods for both indoor and outdoor scenes. We find that our method improves geometry by completing holes (see the chair in the first row), removing floaters (see in front of century plant in the second row and garbage truck in the fourth row), and sharping geometry (see the under the bench in the third row).

\textbf{Ablations of our 3D prior on real data} We ablate our method on the ``garbage'' scene (\cref{tab:nerf_ablation_study}). We find that the cube sampling strategies (i.e., where to apply the diffusion prior) are important, and using the proposed importance sampling with densities yields the best performance. \cref{fig:ablations} compares uniform sampling with importance sampling (using densities). Importance sampling samples less empty space, and thus is more effective at cleaning up floaters and scene geometry. We compare the proposed DSDS loss against SDS with either a scaled and shifted sigmoid or a clamped sigmoid that satisfies our requirements (see \cref{sec:dsds}). We find the gradients do not flow well through this activation function resulting in a distorted scene with many floaters (see \cref{fig:ablations} left). We also ablate the cube sizes used cubes size ranging from $1\%$ to $20\%$ of the scene scale. We find that our method is relatively robust to the cube sizes, yielding a trade-off between removing more with larger cubes and removing less with smaller cubes.

\begin{figure}[t]
    \centering
    \includegraphics[width=1\linewidth]{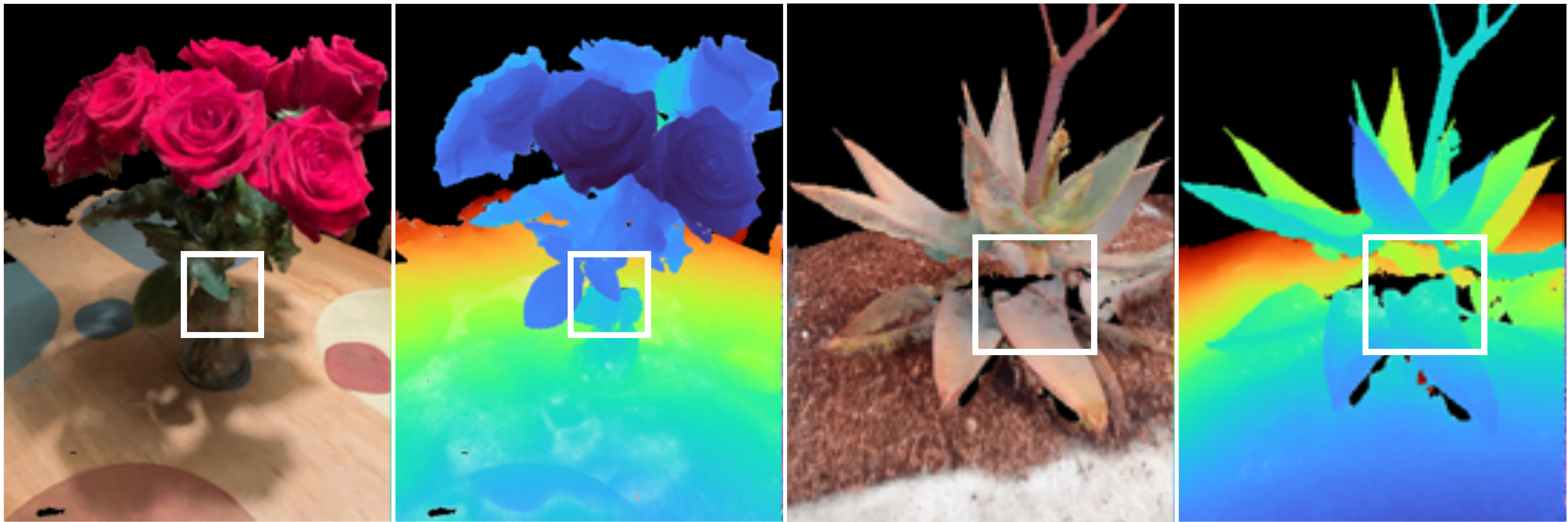}
    \caption{\textbf{Limitations.} The proposed model only operates on densities, which comes with some limitations. We find that it cannot distinguish floaters from transparent objects (left). It does not hallucinate texture and thus ends up removing regions that are occluded in all training views (right).}
    \label{fig:limitations}
\VSPACEAMOUNT
\end{figure}
\vspace{-0.5em}
\section{Conclusion and future work}

\textbf{Transparent objects.} NeRFs are able to represent transparent objects by assigning low densities to the transparent object. These transparent densities behave similarly to floaters, and it requires semantic information to distinguish the two. Since our local diffusion prior does not have semantic information, it removes transparent objects as illustrated in the vase in \cref{fig:limitations}. 

\textbf{Hallucinating texture.} The proposed method cleans geometry but cannot edit texture, as our method operates on densities. This means that we can remove regions that contain floaters or fill holes, but we cannot colorize these regions.
We leave colorization and inpainting low-confidence regions to future work, where 2D diffusion priors~\cite{poole2022dreamfusion,melas2023realfusion} or 3D-consistent inpainting~\cite{liu2021infinite,li2022infinitenature} may be relevant.

\textbf{Conclusion.} We propose a new evaluation procedure of Neural Radiance Fields (NeRFs) that better encompasses how artists, designers, or hobbyists use the technology. We present a dataset with $12$ captures recorded with two camera trajectories each, one used for training and one for evaluation. We find that current hand-crafted regularizers are insufficient when NeRFs are rendered away from the training trajectory. We propose a data-driven, local 3D diffusion prior, Nerfbusters, that removes floaters and improves the scene geometry. We have implemented our proposed evaluation procedure and method in the widely adopted codebase Nerfstudio and will release it for the benefit of the community.

\section*{Acknowledgements}

This project was supported in part by BAIR/BDD. We thank our colleagues for their discussions and feedback throughout the project, especially those within the Kanazawa AI Research (KAIR) lab and those part of the Nerfstudio development team. We thank Kamyar Salahi, Abhik Ahuja, Jake Austin, Xinyang Han, and Alexander Kristoffersen for helping to improve paper drafts.

\appendix
\section{Supplement}



In this appendix, we seek to provide more details about the proposed dataset, evaluation protocol, and per-scene results.

\section{Dataset}\label{sec:appendix_datasets}

We compare the proposed dataset with popular NeRF datasets. We first normalize each scene's camera poses (train and evaluation cameras combined) to the range [-1, 1]. We then find the nearest training image for each evaluation image. We compute the translation and rotation difference between the nearest neighbors (NN) and average over the number of evaluation poses. \cref{tab:related_datset} shows that the proposed Nerfbusters dataset has a larger translation and rotation difference between training and evaluation frames than existing popular NeRF datasets.

\begin{table}[]
    \centering
    \begin{tabular}{l|cc} \toprule
                    & Translation to NN & Rotation to NN \\ \midrule
        Nerfbusters	& 0.62	& 28.51 \\
        Synthetic \cite{mildenhall2021nerf}	& 0.29	& 18.12 \\
        LLFF \cite{mildenhall2019llff}	    & 0.24	& 2.07 \\
        Phototourism \cite{jin2021image}	& 0.01	& 9.47 \\
        MipNeRF-360 \cite{barron2022mip}	& 0.07	& 10.49 \\ \bottomrule
    \end{tabular}
    \caption{\textbf{Distance between training and eval images.} Average translation and rotation difference between eval poses and closest training poses (nearest neighbor (NN)). The proposed Nerfbusters dataset is larger in translation and rotation differences than the current popular NeRF datasets.}
    \label{tab:related_datset}
\end{table}

\cref{tab:related_datset} compares translation and rotation differences between training and evaluation images. Although this is a useful approximation of the ``difficulty" of the datasets, it does not encompass variations in the camera intrinsics, e.g. we can obtain a similar effect as a translation by changing the focal length. In practice though, a common assumption is that the same camera intrinsics are used for training and test renderings. This assumption does not hold for the Phototourism \cite{jin2021image} dataset, where the camera intrinsics vary for each training frame. Therefore, \cref{tab:related_datset} is a crude approximation of difficulty for this dataset. 

\begin{figure*}
     \centering
     \begin{subfigure}[b]{0.19\textwidth}
         \centering
         \caption{Drum}
         \includegraphics[width=\textwidth]{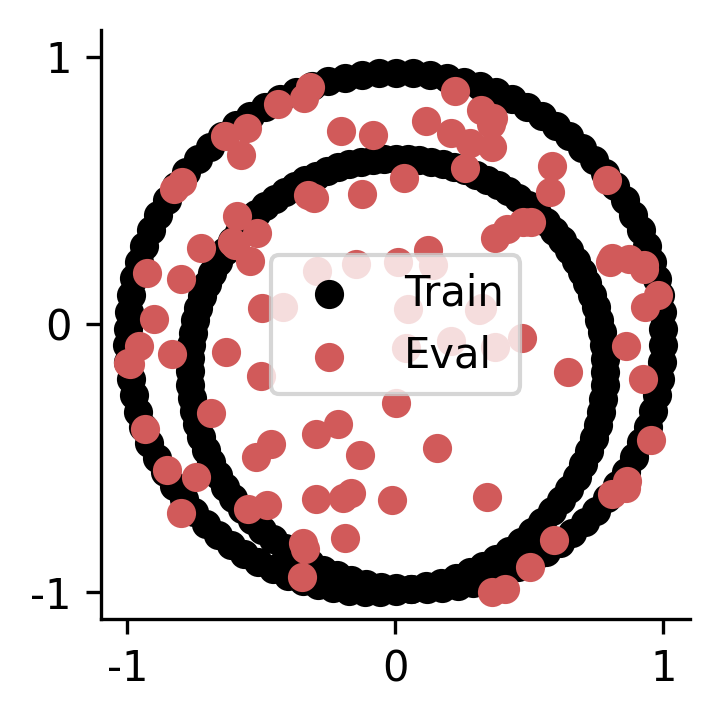}
         
     \end{subfigure}
     \begin{subfigure}[b]{0.19\textwidth}
         \centering
         \caption{Ficus}
         \includegraphics[width=\textwidth]{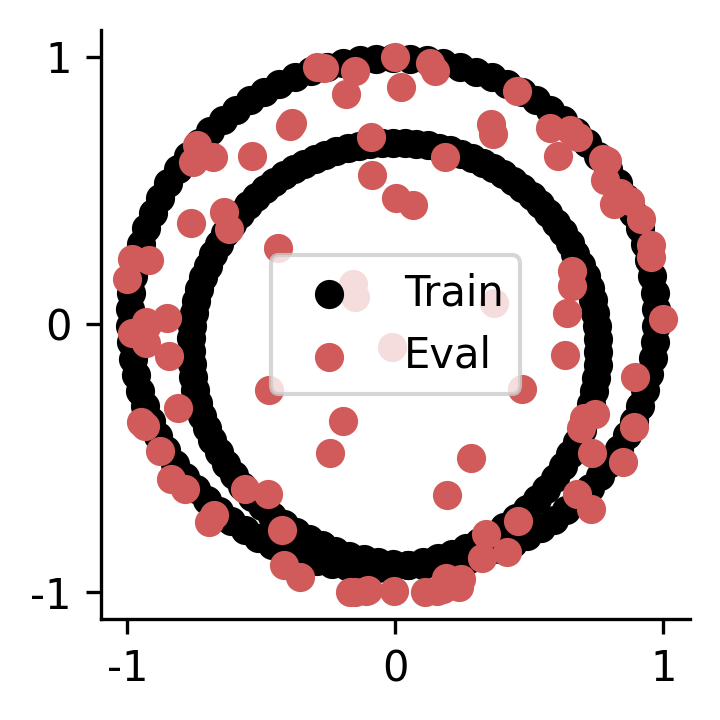}
     \end{subfigure}
     \begin{subfigure}[b]{0.19\textwidth}
         \centering
         \caption{Hotdog}
         \includegraphics[width=\textwidth]{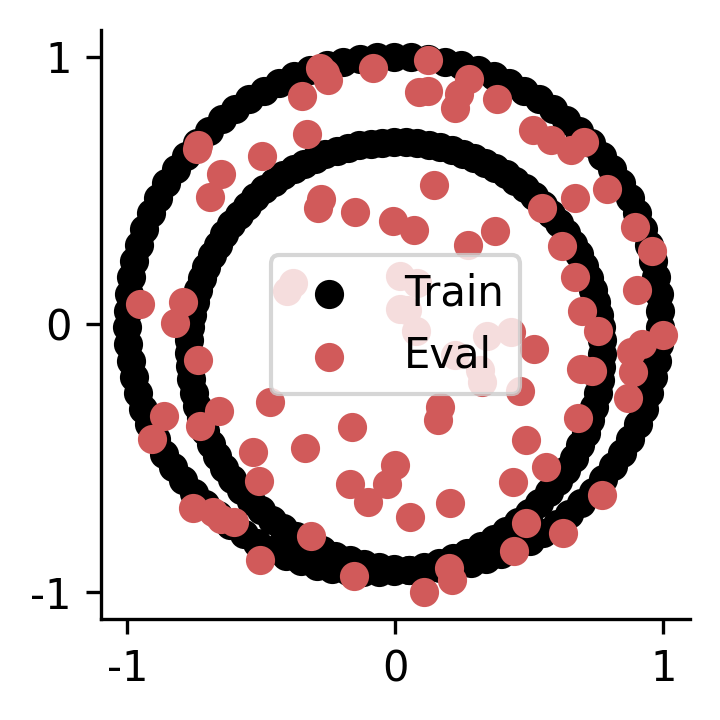}
         
     \end{subfigure}
     \begin{subfigure}[b]{0.19\textwidth}
         \centering
         \caption{Lego}
         \includegraphics[width=\textwidth]{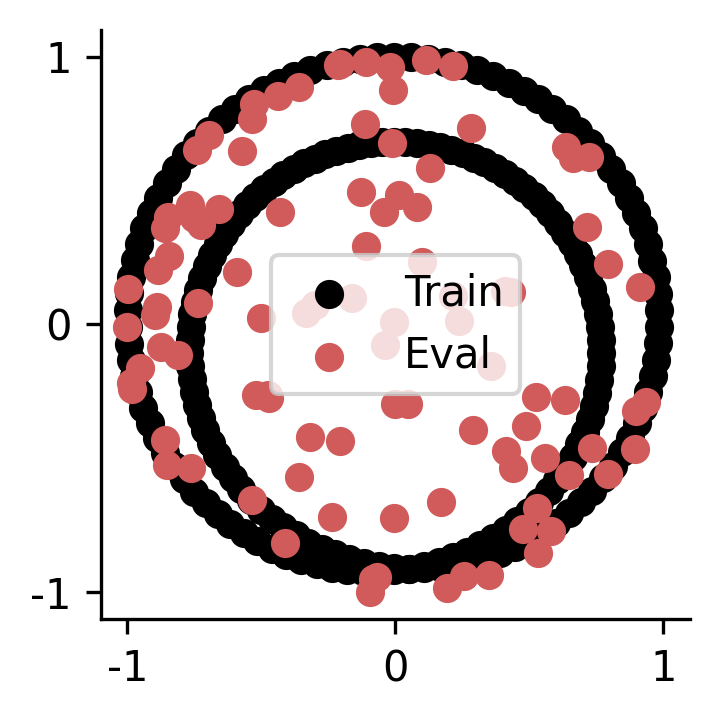}
         
     \end{subfigure}
     \begin{subfigure}[b]{0.19\textwidth}
         \centering
         \caption{Materials}
         \includegraphics[width=\textwidth]{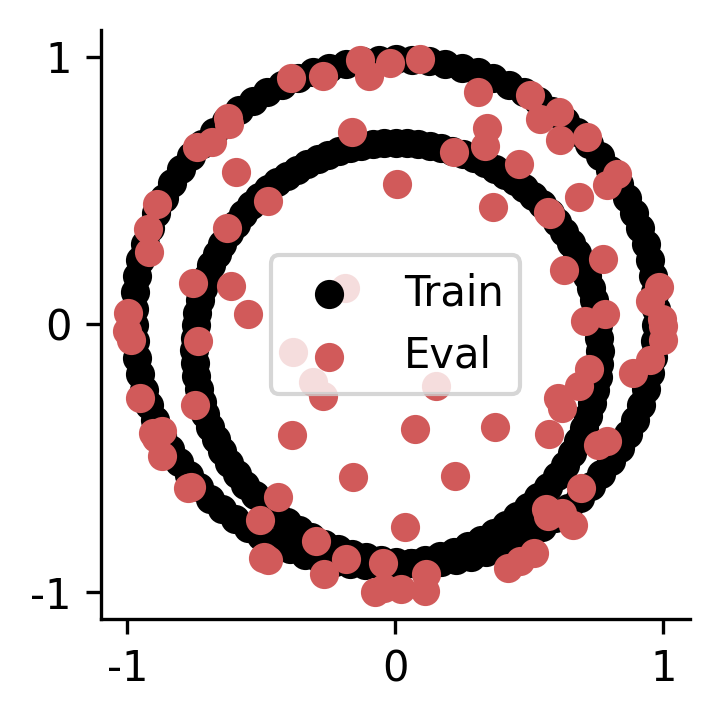}
         
     \end{subfigure}
     \begin{subfigure}[b]{0.19\textwidth}
         \centering
         \caption{Mic}
         \includegraphics[width=\textwidth]{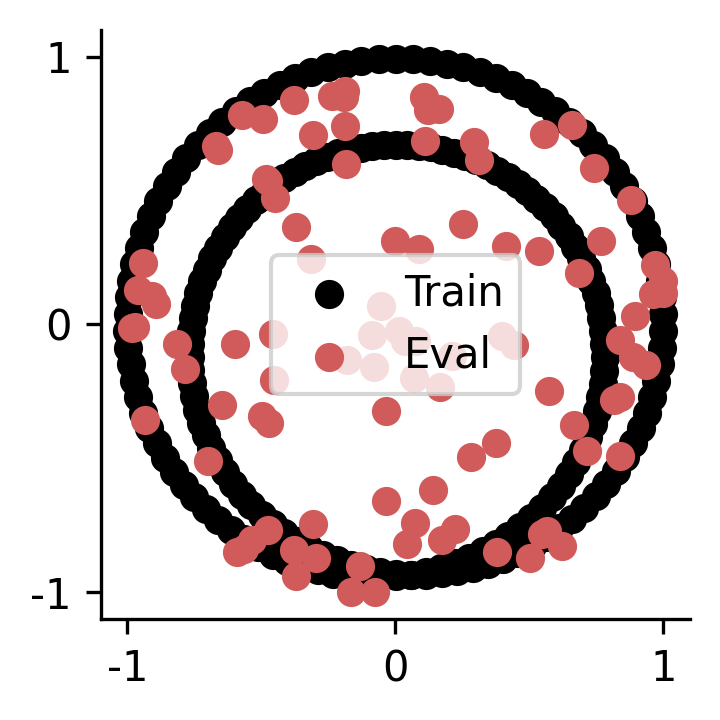}
         
     \end{subfigure}
     \begin{subfigure}[b]{0.19\textwidth}
         \centering
         \caption{Ship}
         \includegraphics[width=\textwidth]{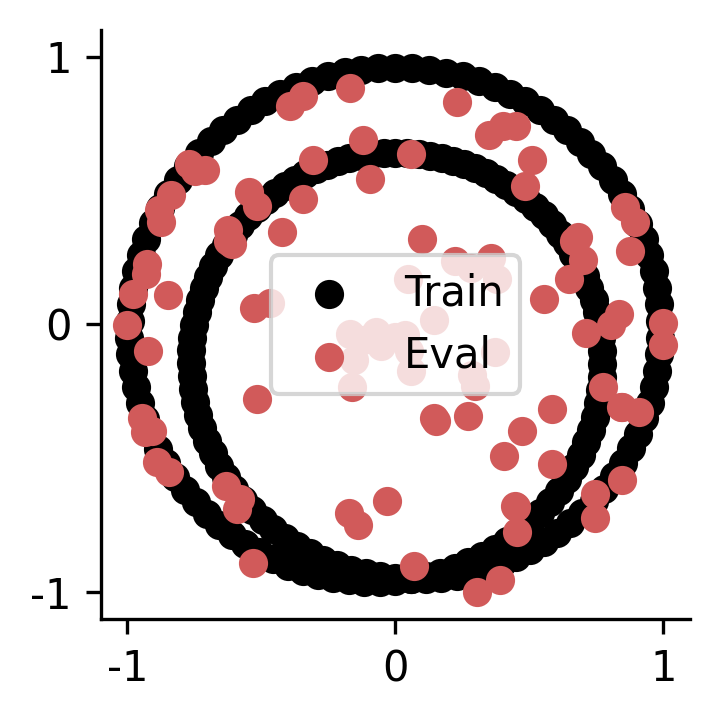}
         
     \end{subfigure}
        \caption{\textbf{Synethic NeRF.} Birds-eye view of the training and evaluation cameras for each scene in the dataset. Each point represents the xy projection of the camera locations. The black dots are cameras used for training and the red dots show views for evaluation.}
        \label{fig:synethtic}
\end{figure*}

\begin{figure*}
     \centering
     \begin{subfigure}[b]{0.19\textwidth}
         \centering
         \caption{Fern}
         \includegraphics[width=\textwidth]{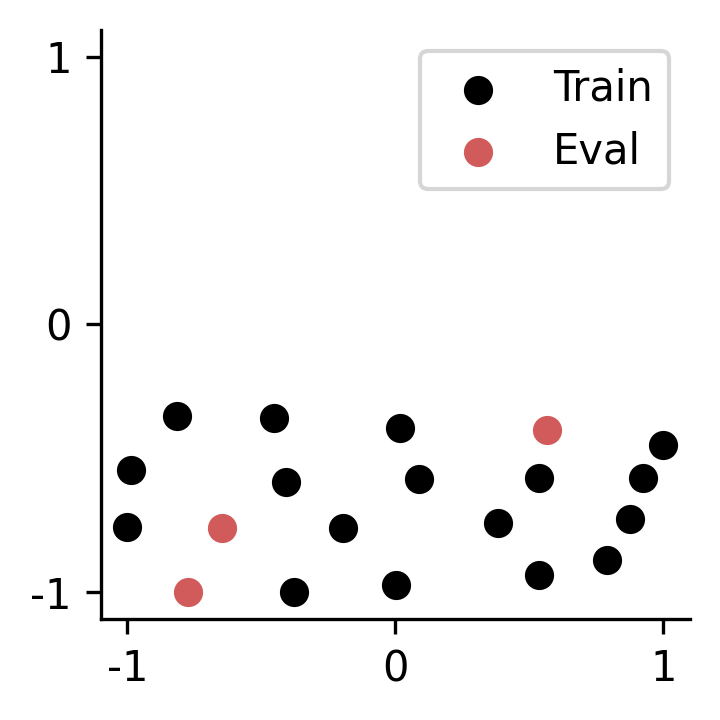}
         
     \end{subfigure}
     \begin{subfigure}[b]{0.19\textwidth}
         \centering
         \caption{Flower}
         \includegraphics[width=\textwidth]{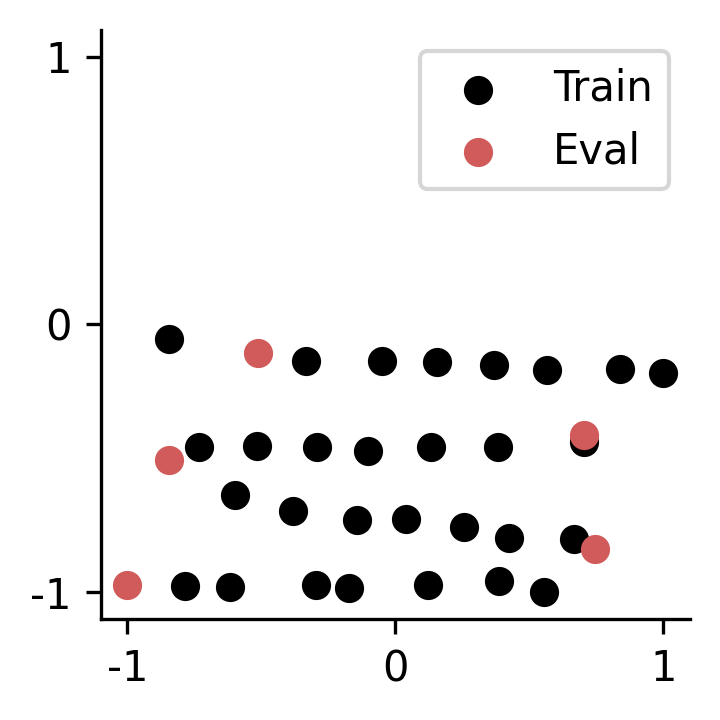}
         
     \end{subfigure}
     \begin{subfigure}[b]{0.19\textwidth}
         \centering
         \caption{Fortress}
         \includegraphics[width=\textwidth]{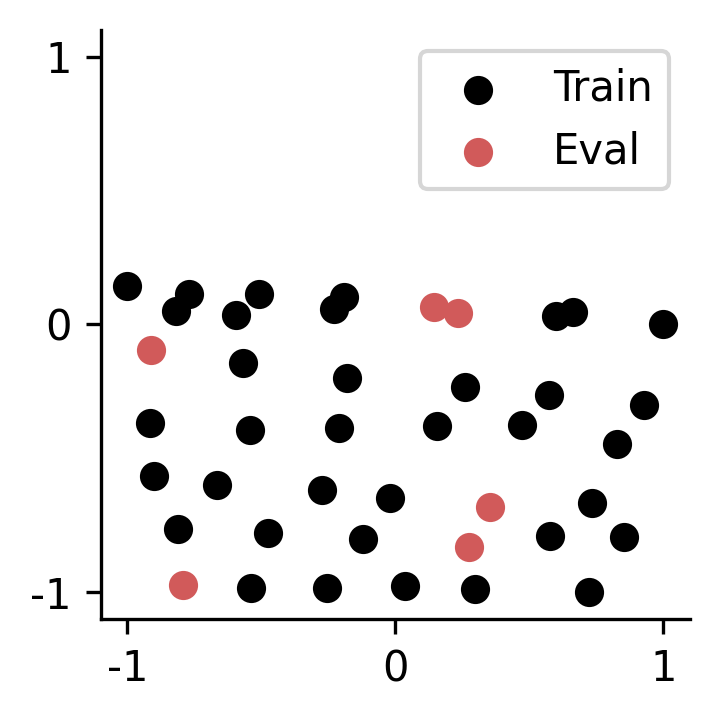}
         
     \end{subfigure}
     \begin{subfigure}[b]{0.19\textwidth}
         \centering
         \caption{Horns}
         \includegraphics[width=\textwidth]{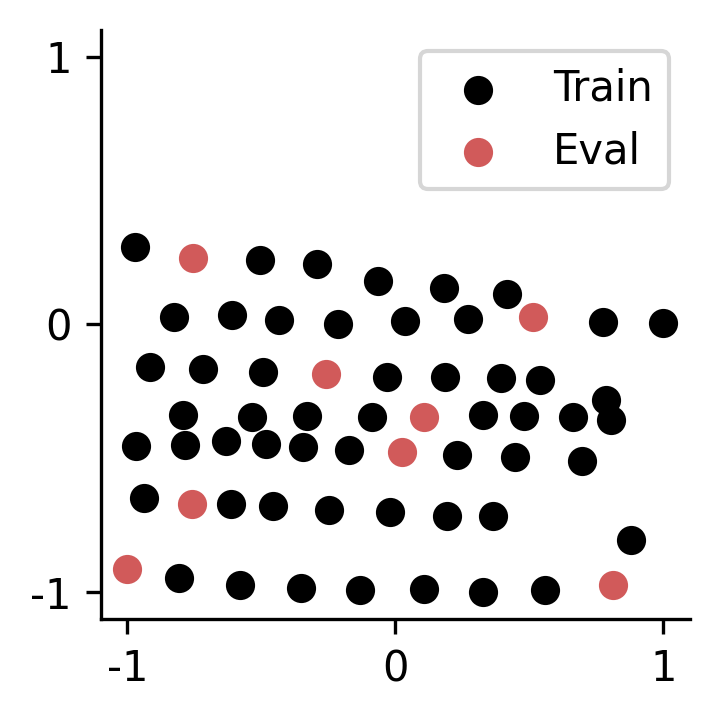}
         
     \end{subfigure}
     \begin{subfigure}[b]{0.19\textwidth}
         \centering
         \caption{Leaves}
         \includegraphics[width=\textwidth]{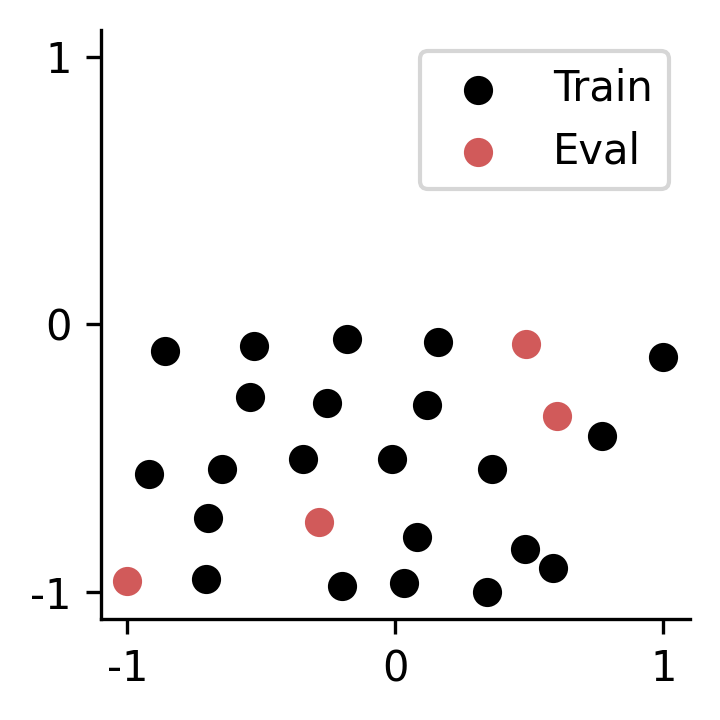}
         
     \end{subfigure}
     \begin{subfigure}[b]{0.19\textwidth}
         \centering
         \caption{Orchids}
         \includegraphics[width=\textwidth]{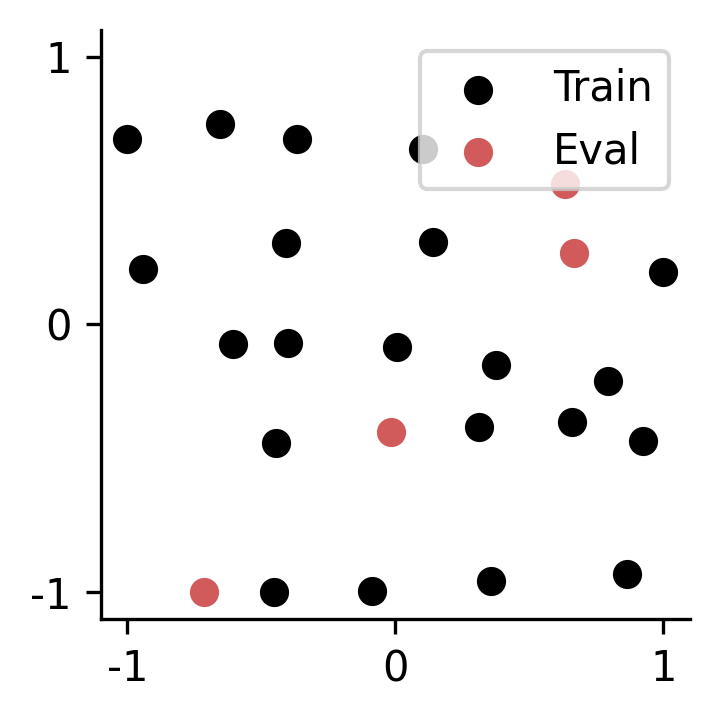}
         
     \end{subfigure}
     \begin{subfigure}[b]{0.19\textwidth}
         \centering
         \caption{Room}
         \includegraphics[width=\textwidth]{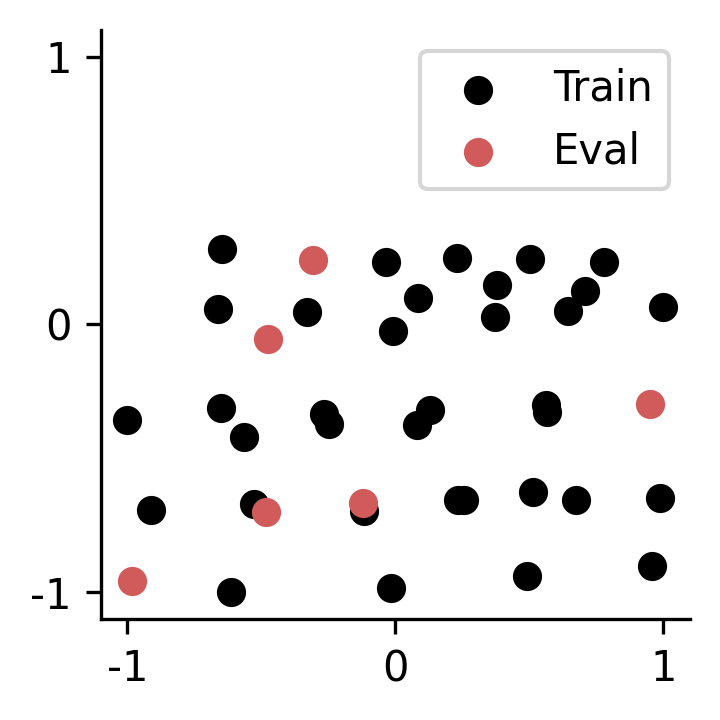}
         
     \end{subfigure}
     \begin{subfigure}[b]{0.19\textwidth}
         \centering
         \caption{Trex}
         \includegraphics[width=\textwidth]{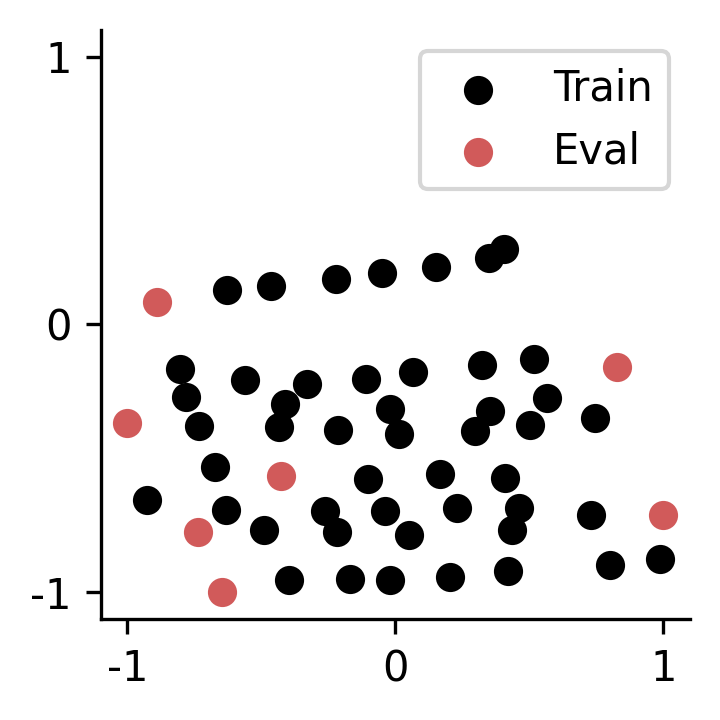}
         
     \end{subfigure}
        \caption{\textbf{LLFF dataset.} Birds-eye view of the training and evaluation cameras for each scene in the dataset. Each point represents the xy projection of the camera locations. The black dots are cameras used for training and the red dots show views for evaluation.}
        \label{fig:llff}
\end{figure*}

\begin{figure*}
     \centering
     \begin{subfigure}[b]{0.19\textwidth}
         \centering
         \caption{Room}
         \includegraphics[width=\textwidth]{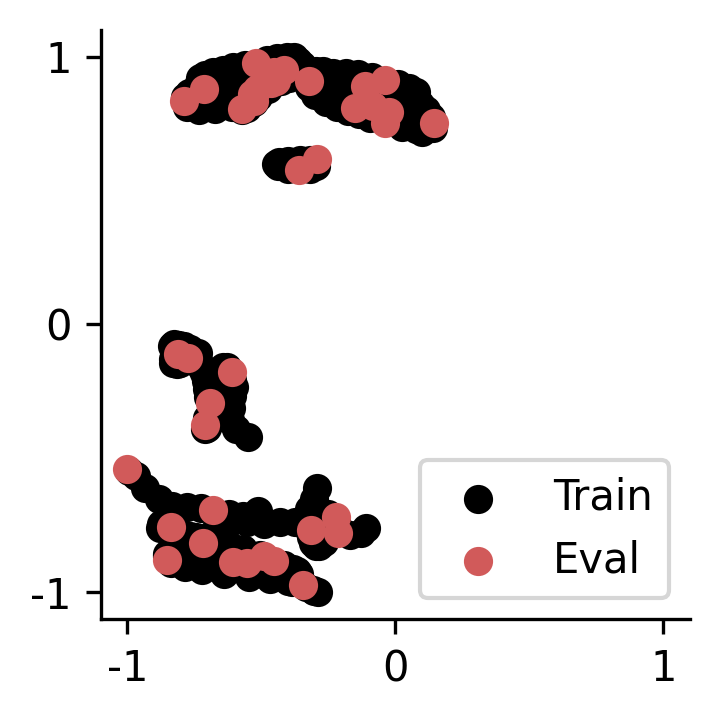}
         
     \end{subfigure}
     \begin{subfigure}[b]{0.19\textwidth}
         \centering
         \caption{Bicycle}
         \includegraphics[width=\textwidth]{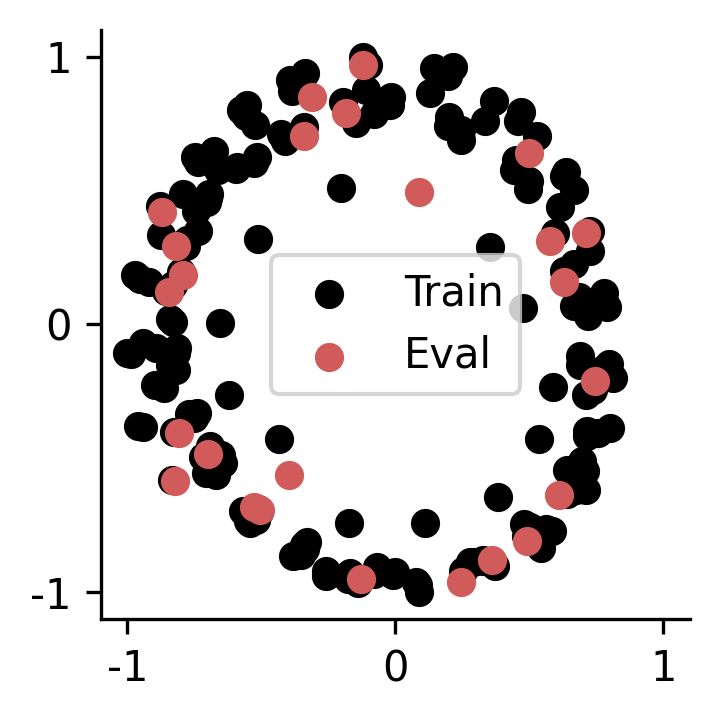}
         
     \end{subfigure}
     \begin{subfigure}[b]{0.19\textwidth}
         \centering
         \caption{Bonsai}
         \includegraphics[width=\textwidth]{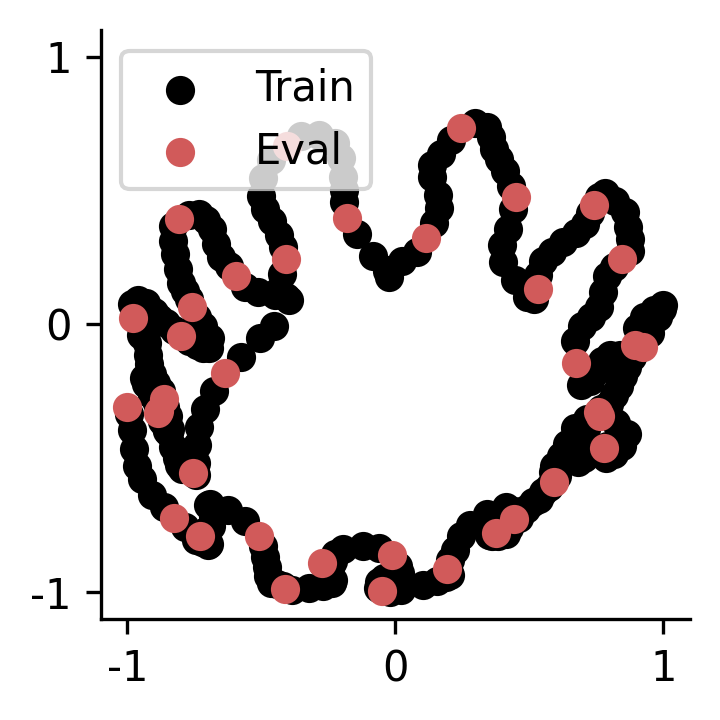}
         
     \end{subfigure}
     \begin{subfigure}[b]{0.19\textwidth}
         \centering
         \caption{Counter}
         \includegraphics[width=\textwidth]{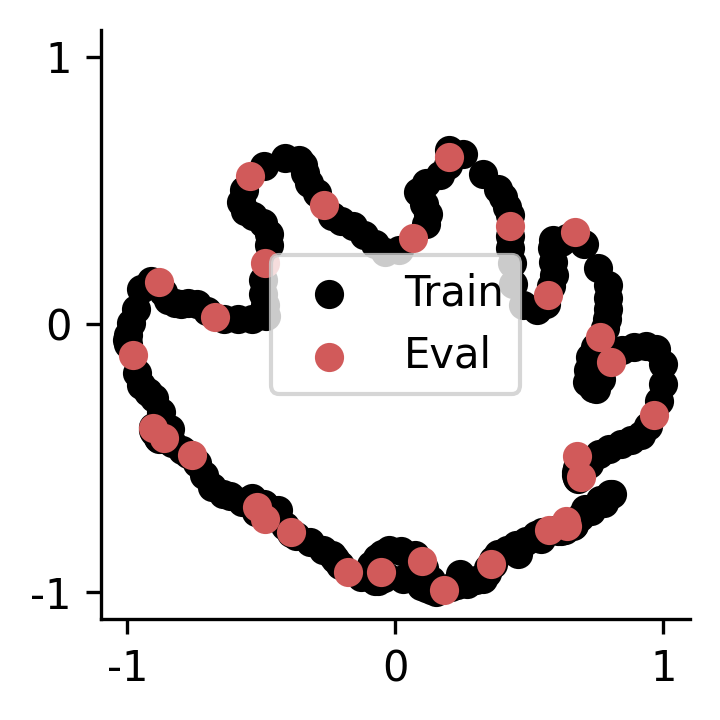}
         
     \end{subfigure}
     \begin{subfigure}[b]{0.19\textwidth}
         \centering
         \caption{Garden}
         \includegraphics[width=\textwidth]{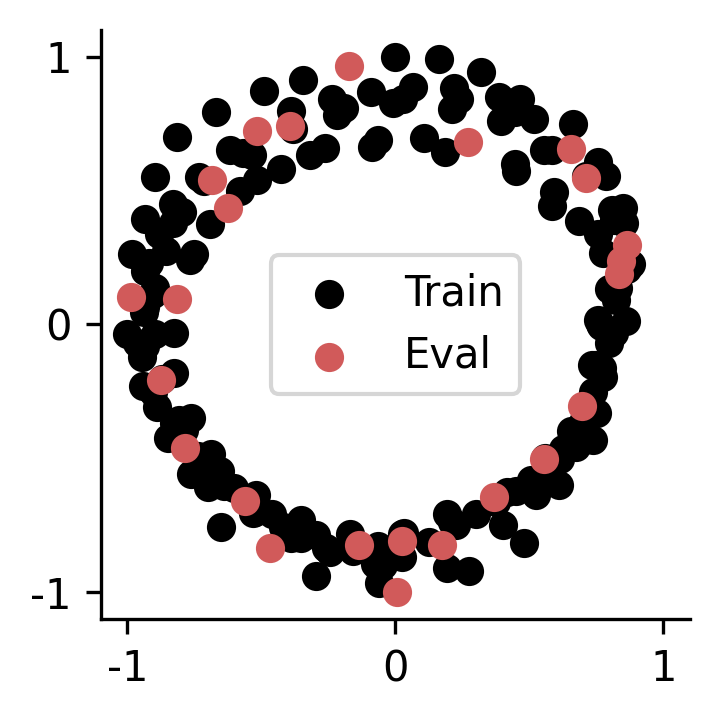}
         
     \end{subfigure}
     \begin{subfigure}[b]{0.19\textwidth}
         \centering
         \caption{Stump}
         \includegraphics[width=\textwidth]{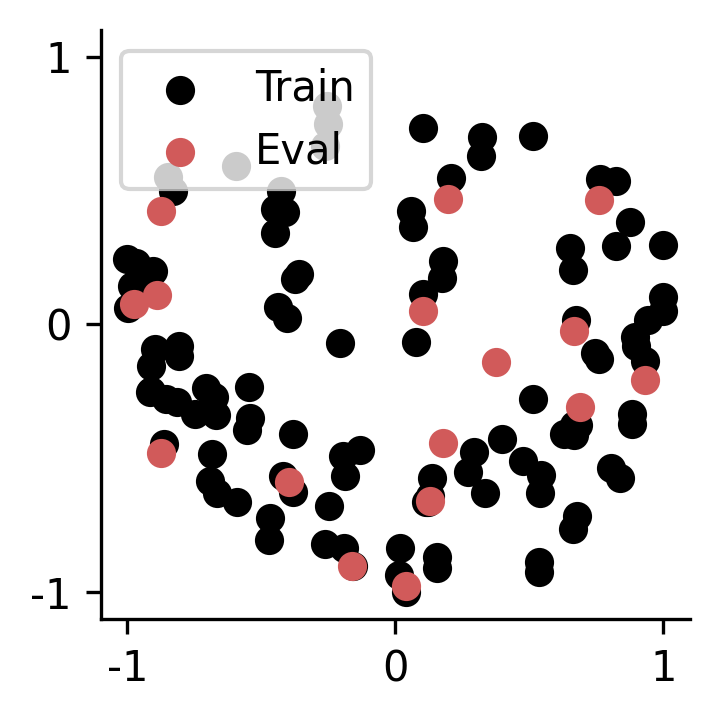}
         
     \end{subfigure}
        \caption{\textbf{MipNeRF 360 dataset.} Birds-eye view of the training and evaluation cameras for each scene in the dataset. Each point represents the xy projection of the camera locations. The black dots are cameras used for training and the red dots show views for evaluation.}
        \label{fig:mip360}
\end{figure*}

\begin{figure*}
     \centering
     \begin{subfigure}[b]{0.19\textwidth}
         \centering
         \caption{Brandenburg}
         \includegraphics[width=\textwidth]{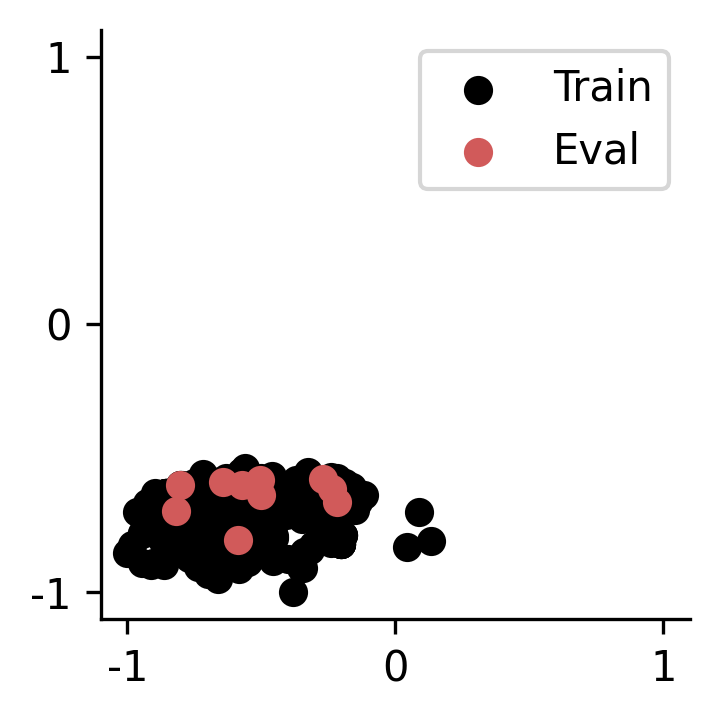}
         
     \end{subfigure}
     \begin{subfigure}[b]{0.19\textwidth}
         \centering
         \caption{Trevi}
         \includegraphics[width=\textwidth]{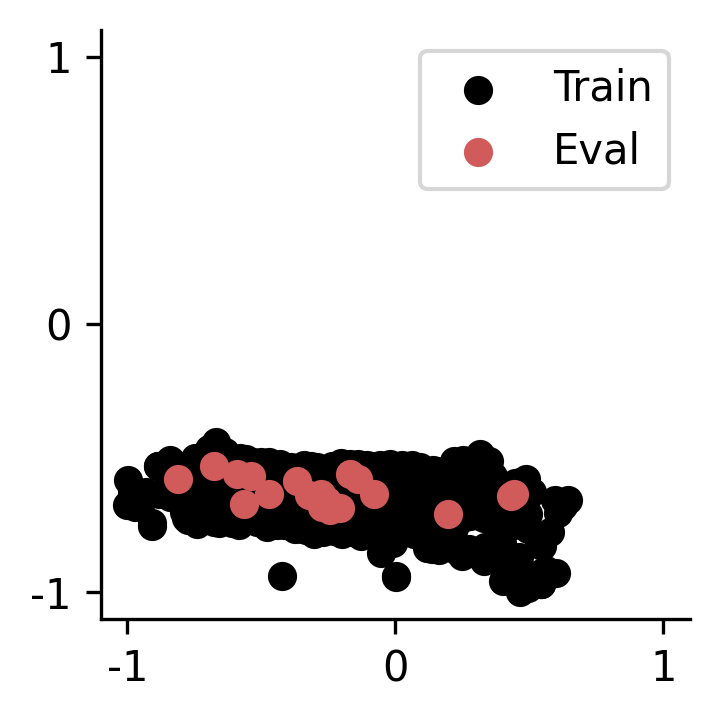}
         
     \end{subfigure}
     \begin{subfigure}[b]{0.19\textwidth}
         \centering
         \caption{Sacre}
         \includegraphics[width=\textwidth]{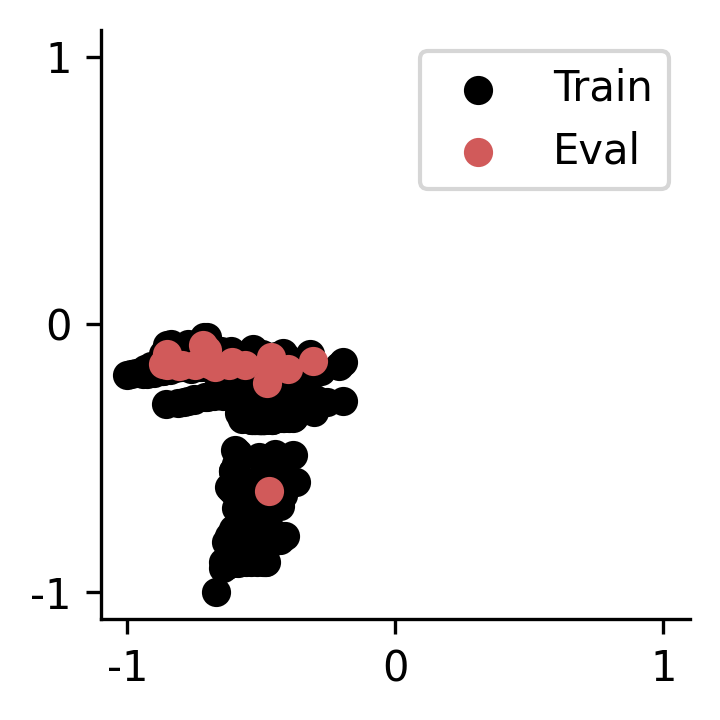}
         
     \end{subfigure}
        \caption{\textbf{Phototourism dataset.} Birds-eye view of the training and evaluation cameras for each scene in the dataset. Each point represents the xy projection of the camera locations. The black dots are cameras used for training and the red dots show views for evaluation.}
        \label{fig:photo}
\end{figure*}

\begin{figure*}
     \centering
     \begin{subfigure}[b]{0.19\textwidth}
         \centering
         \caption{Aloe}
         \includegraphics[width=\textwidth]{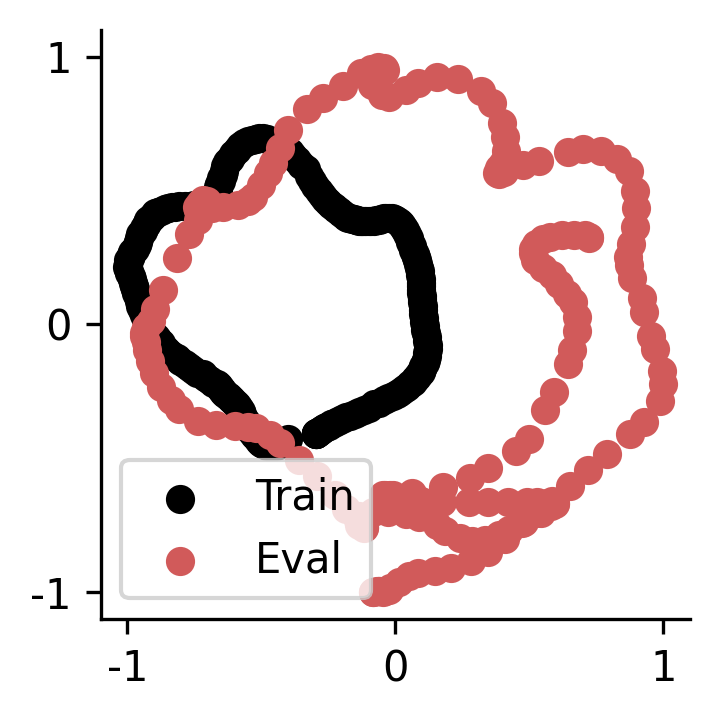}
         
     \end{subfigure}
     \begin{subfigure}[b]{0.19\textwidth}
         \centering
         \caption{Art}
         \includegraphics[width=\textwidth]{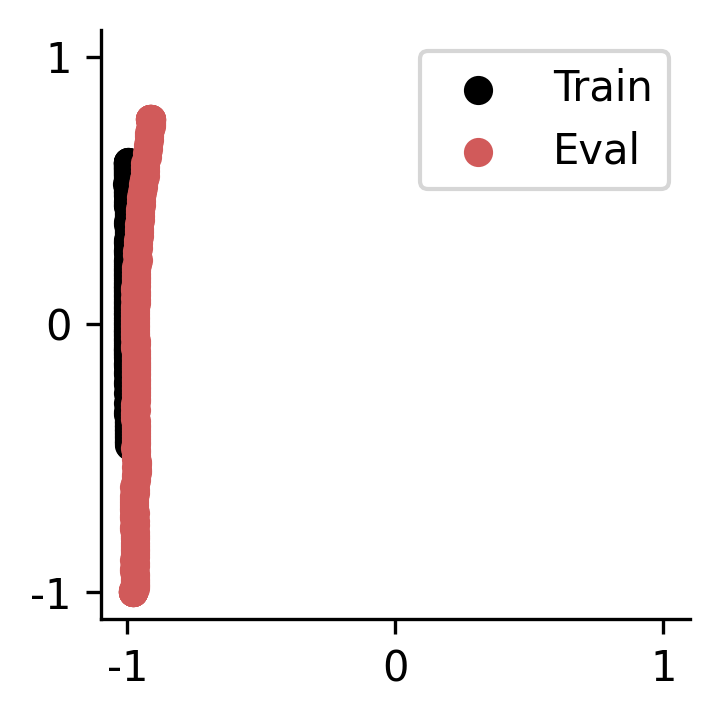}
         
     \end{subfigure}
     \begin{subfigure}[b]{0.19\textwidth}
         \centering
         \caption{Car}
         \includegraphics[width=\textwidth]{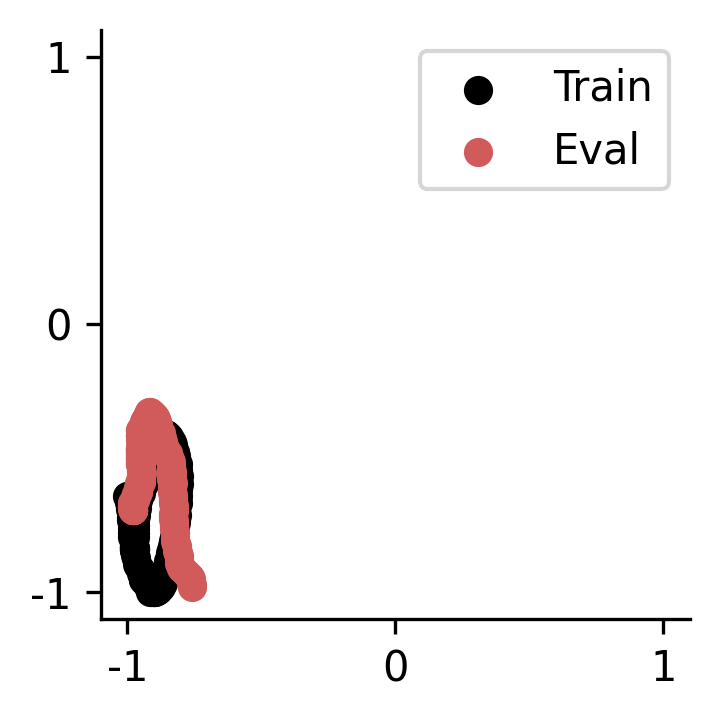}
         
     \end{subfigure}
     \begin{subfigure}[b]{0.19\textwidth}
         \centering
         \caption{Century}
         \includegraphics[width=\textwidth]{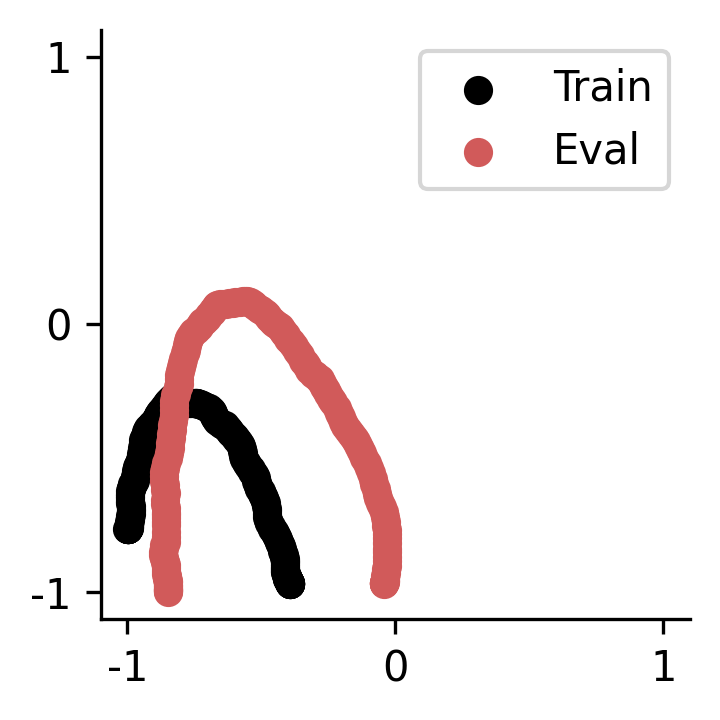}
         
     \end{subfigure}
     \begin{subfigure}[b]{0.19\textwidth}
         \centering
         \caption{Flowers}
         \includegraphics[width=\textwidth]{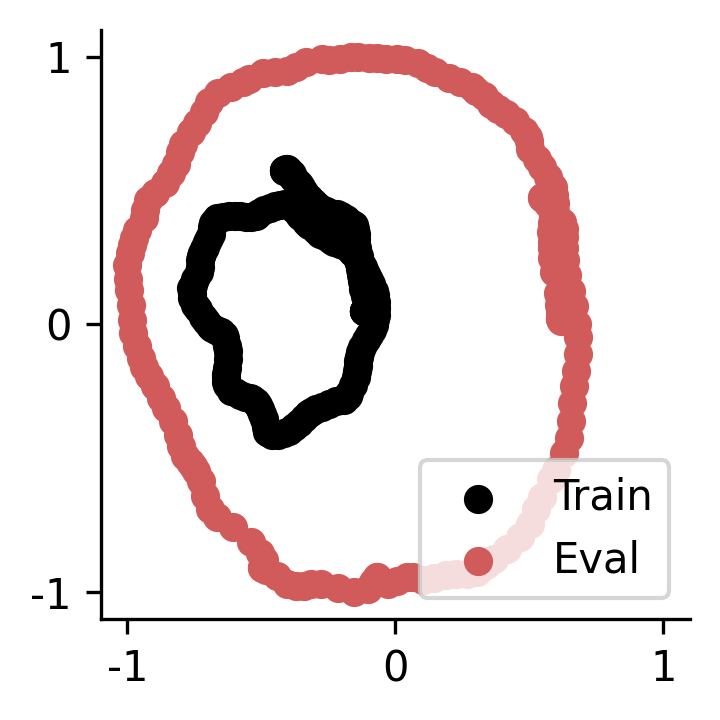}
         
     \end{subfigure}
     \begin{subfigure}[b]{0.19\textwidth}
         \centering
         \caption{Garbage}
         \includegraphics[width=\textwidth]{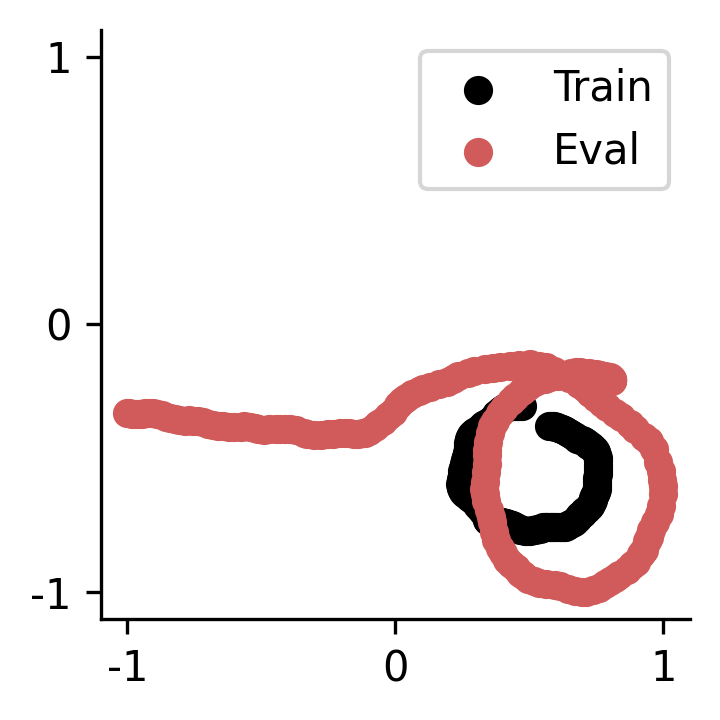}
         
     \end{subfigure}
     \begin{subfigure}[b]{0.19\textwidth}
         \centering
         \caption{Picnic}
         \includegraphics[width=\textwidth]{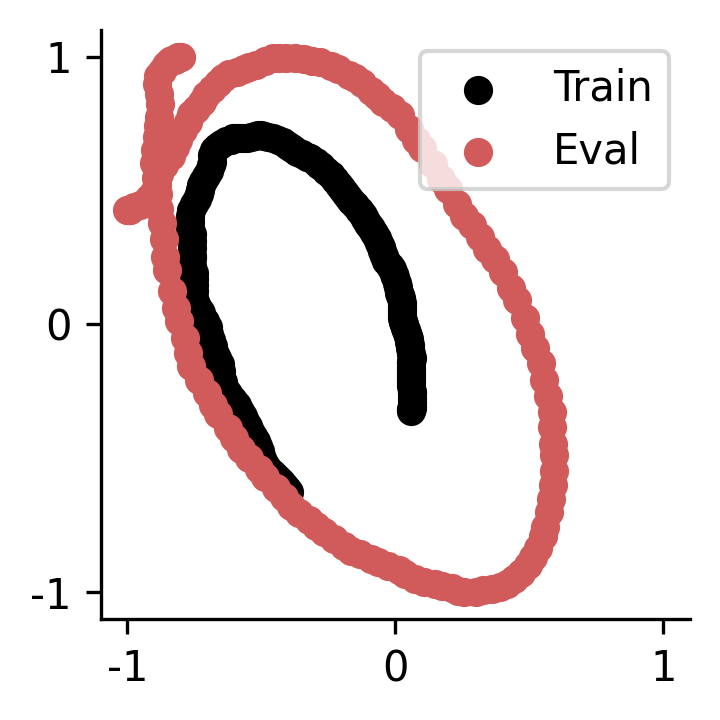}
         
     \end{subfigure}
     \begin{subfigure}[b]{0.19\textwidth}
         \centering
         \caption{Pikachu}
         \includegraphics[width=\textwidth]{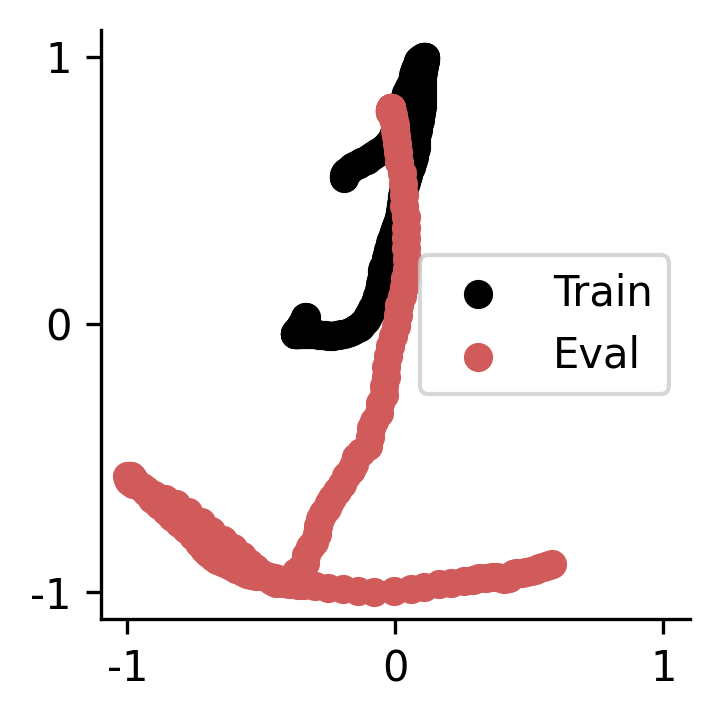}
         
     \end{subfigure}
     \begin{subfigure}[b]{0.19\textwidth}
         \centering
         \caption{Pipe}
         \includegraphics[width=\textwidth]{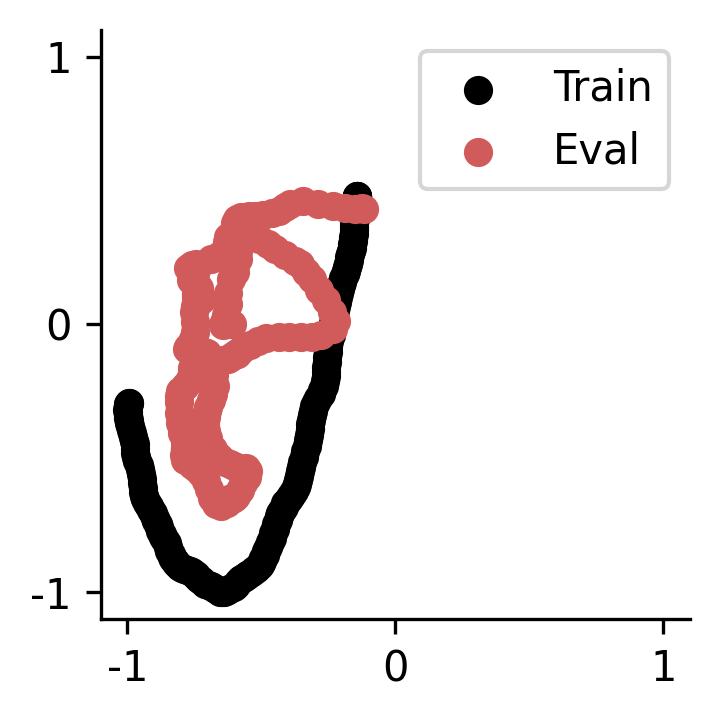}
         
     \end{subfigure}
     \begin{subfigure}[b]{0.19\textwidth}
         \centering
         \caption{Roses}
         \includegraphics[width=\textwidth]{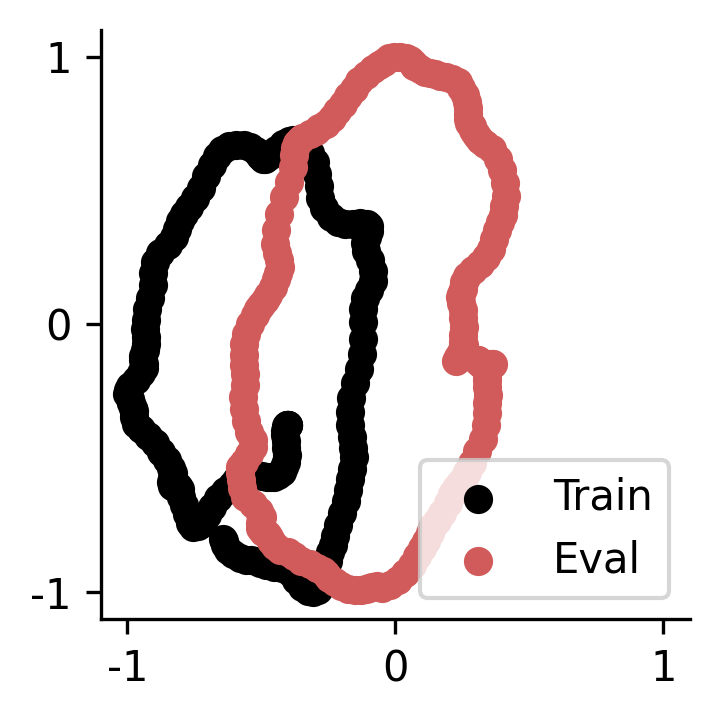}
         
     \end{subfigure}
     \begin{subfigure}[b]{0.19\textwidth}
         \centering
         \caption{Stairs}
         \includegraphics[width=\textwidth]{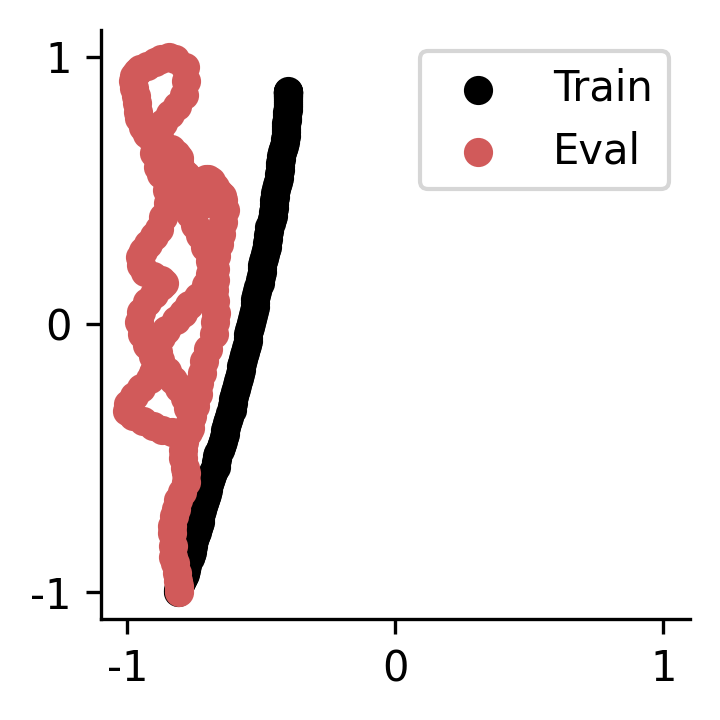}
         
     \end{subfigure}
     \begin{subfigure}[b]{0.19\textwidth}
         \centering
         \caption{Table}
         \includegraphics[width=\textwidth]{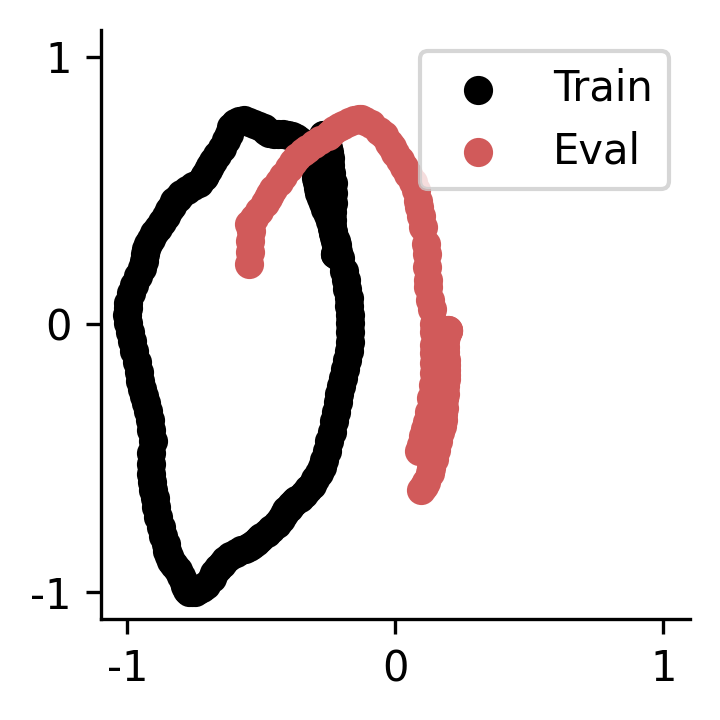}
         
     \end{subfigure}
        \caption{\textbf{Nerfbusters dataset.} Birds-eye view of the training and evaluation cameras for each scene in the dataset. Each point represents the xy projection of the camera locations. The black dots are cameras used for training and the red dots show views for evaluation.}
        \label{fig:cleanerf}
\end{figure*}

\cref{fig:synethtic,fig:llff,fig:mip360,fig:photo,fig:cleanerf} show top-down views of training and evaluation views for each scene in the Synthetic \cite{mildenhall2021nerf}, LLFF \cite{mildenhall2019llff}, MipNeRF-360 \cite{barron2022mip}, Phototourism \cite{jin2021image}, and our Nerfbusters dataset. Most of the datasets use every 8th frame for evaluation, resulting in training and evaluation views being very close. The train and test frames are further apart in the Nerfbusters dataset, faithfully representing some of the real-world challenges that users of NeRF experience when rendering fly-throughs of their captured scenes.


\section{Per-Scene Evaluation}\label{sec:appendix_results}

We provide per-scene evaluation for our method and the baseline methods in \cref{tab:aloe,tab:art,tab:car,tab:century,tab:flowers,tab:garbage,tab:picnic,tab:pikachu,tab:pipe,tab:plant,tab:roses,tab:table}. The average over the 12 datasets is included in the main paper.

\begin{table*}[]
    \centering
    \begin{tabular}{l|lll|ll|lllll|l} 
        \toprule
        
 & PSNR $\uparrow$ & SSIM $\uparrow$ & LPIPS $\downarrow$ & Depth $\downarrow$ & Disp. $\downarrow$ & Mean $^{\circ}$ $\downarrow$ & Median $^{\circ}$ $\downarrow$ & \% $30^{\circ}$ $\uparrow$ & Coverage $\uparrow$ \\
        \midrule
        
 Nerfacto Pseudo GT    & 24.02 & 0.8591 & 0.0801 & 0.00 & 0.00 & 0.01 & 0.00 & 1.00 & 0.84 \\ \midrule
 Nerfacto              & 18.33 & 0.6037 & 0.2968 & 0.01 & 0.04 & 45.50 & 39.01 & 0.37 & 0.84 \\ 
  + Visibility Loss    & 18.32 & 0.6063 & 0.2929 & 0.01 & 0.02 & 45.28 & 38.81 & 0.37 & 0.84 \\ 
  + Vis + Sparsity     & 18.41 & 0.6082 & 0.2895 & 0.01 & 0.02 & 45.19 & 38.76 & 0.37 & 0.84 \\ 
  + Vis + TV           & 18.31 & 0.6073 & 0.2983 & 0.01 & 0.02 & 46.12 & 39.34 & 0.36 & 0.84 \\ 
  + Vis + RegNeRF      & 18.31 & 0.6081 & 0.2933 & 0.01 & 0.02 & 45.44 & 39.04 & 0.37 & 0.84 \\ 
  + Vis + DSDS (Ours)  & 18.75 & 0.6390 & 0.2441 & 0.01 & 0.02 & 45.43 & 39.14 & 0.36 & 0.80 \\ 

        \bottomrule
    \end{tabular}
    \caption{\textbf{``aloe" capture quantitative evaluation.} Per scene results that compare Nerfacto with baselines (that uses geometric handcrafted regularizers) and Nerfbusters (that uses a data-driven local 3D prior). We highlight that all baselines use the proposed visibility loss. \cref{sec:evaluation_procedure} describes metrics in more detail, $\uparrow$/$\downarrow$ indicates if higher/lower is better.}
    \label{tab:aloe}
\end{table*}

\begin{table*}[]
    \centering
    \begin{tabular}{l|lll|ll|lllll|l} 
        \toprule
        
 & PSNR $\uparrow$ & SSIM $\uparrow$ & LPIPS $\downarrow$ & Depth $\downarrow$ & Disp. $\downarrow$ & Mean $^{\circ}$ $\downarrow$ & Median $^{\circ}$ $\downarrow$ & \% $30^{\circ}$ $\uparrow$ & Coverage $\uparrow$ \\
        \midrule
        
 Nerfacto Pseudo GT    & 25.25 & 0.8534 & 0.0710 & 0.00 & 0.00 & 0.01 & 0.00 & 1.00 & 0.88 \\ \midrule
 Nerfacto              & 14.72 & 0.3784 & 0.3978 & 1025.94 & 0.20 & 65.26 & 60.02 & 0.20 & 0.87 \\ 
  + Visibility Loss    & 14.77 & 0.3860 & 0.3933 & 1006.97 & 0.17 & 64.55 & 59.30 & 0.20 & 0.86 \\ 
  + Vis + Sparsity     & 14.86 & 0.3900 & 0.3901 & 907.83 & 0.17 & 64.48 & 59.19 & 0.20 & 0.86 \\ 
  + Vis + TV           & 15.02 & 0.3991 & 0.3838 & 719.83 & 0.16 & 68.80 & 63.99 & 0.17 & 0.85 \\ 
  + Vis + RegNeRF      & 13.30 & 0.2995 & 0.4647 & 1386.03 & 2.57 & 72.13 & 67.93 & 0.15 & 0.89 \\ 
  + Vis + DSDS (Ours)  & 15.05 & 0.4336 & 0.3408 & 390.80 & 0.14 & 62.21 & 56.89 & 0.21 & 0.74 \\ 

        \bottomrule
    \end{tabular}
    \caption{\textbf{``art" capture quantitative evaluation.} Per scene results that compare Nerfacto with baselines (that uses geometric handcrafted regularizers) and Nerfbusters (that uses a data-driven local 3D prior). We highlight that all baselines use the proposed visibility loss. \cref{sec:evaluation_procedure} describes metrics in more detail, $\uparrow$/$\downarrow$ indicates if higher/lower is better.}
    \label{tab:art}
\end{table*}

\begin{table*}[]
    \centering
    \begin{tabular}{l|lll|ll|lllll|l} 
        \toprule
        
 & PSNR $\uparrow$ & SSIM $\uparrow$ & LPIPS $\downarrow$ & Depth $\downarrow$ & Disp. $\downarrow$ & Mean $^{\circ}$ $\downarrow$ & Median $^{\circ}$ $\downarrow$ & \% $30^{\circ}$ $\uparrow$ & Coverage $\uparrow$ \\
        \midrule
        
 Nerfacto Pseudo GT    & 22.98 & 0.7480 & 0.2093 & 0.00 & 0.00 & 0.01 & 0.00 & 1.00 & 0.80 \\ \midrule
 Nerfacto              & 16.57 & 0.4630 & 0.4090 & 240.77 & 0.56 & 70.51 & 66.31 & 0.17 & 0.81 \\ 
  + Visibility Loss    & 16.67 & 0.4609 & 0.4136 & 133.80 & 0.51 & 70.19 & 65.94 & 0.18 & 0.81 \\ 
  + Vis + Sparsity     & 16.70 & 0.4632 & 0.4149 & 137.97 & 0.38 & 70.57 & 66.38 & 0.17 & 0.81 \\ 
  + Vis + TV           & 16.87 & 0.4827 & 0.3956 & 116.45 & 0.39 & 73.52 & 69.63 & 0.16 & 0.78 \\ 
  + Vis + RegNeRF      & 16.62 & 0.4654 & 0.4099 & 745.14 & 0.56 & 70.73 & 66.59 & 0.17 & 0.82 \\ 
  + Vis + DSDS (Ours)  & 17.40 & 0.5898 & 0.2219 & 23.76 & 0.14 & 65.99 & 60.43 & 0.21 & 0.55 \\ 

        \bottomrule
    \end{tabular}
    \caption{\textbf{``car" capture quantitative evaluation.} Per scene results that compare Nerfacto with baselines (that uses geometric handcrafted regularizers) and Nerfbusters (that uses a data-driven local 3D prior). We highlight that all baselines use the proposed visibility loss. \cref{sec:evaluation_procedure} describes metrics in more detail, $\uparrow$/$\downarrow$ indicates if higher/lower is better.}
    \label{tab:car}
\end{table*}

\begin{table*}[]
    \centering
    \begin{tabular}{l|lll|ll|lllll|l} 
        \toprule
        
 & PSNR $\uparrow$ & SSIM $\uparrow$ & LPIPS $\downarrow$ & Depth $\downarrow$ & Disp. $\downarrow$ & Mean $^{\circ}$ $\downarrow$ & Median $^{\circ}$ $\downarrow$ & \% $30^{\circ}$ $\uparrow$ & Coverage $\uparrow$ \\
        \midrule
        
 Nerfacto Pseudo GT    & 23.76 & 0.8699 & 0.0790 & 0.00 & 0.00 & 0.01 & 0.00 & 1.00 & 0.95 \\ \midrule
 Nerfacto              & 13.97 & 0.3938 & 0.4998 & 2.27 & 3.34 & 65.46 & 59.70 & 0.22 & 0.97 \\ 
  + Visibility Loss    & 13.90 & 0.4022 & 0.4858 & 1.27 & 3.50 & 63.58 & 57.34 & 0.24 & 0.96 \\ 
  + Vis + Sparsity     & 13.91 & 0.4021 & 0.4899 & 1.27 & 2.35 & 63.44 & 57.26 & 0.24 & 0.96 \\ 
  + Vis + TV           & 14.25 & 0.4286 & 0.4464 & 1.03 & 1.40 & 69.38 & 64.59 & 0.18 & 0.96 \\ 
  + Vis + RegNeRF      & 12.63 & 0.3459 & 0.5672 & 2.84 & 3.95 & 71.88 & 67.44 & 0.18 & 0.97 \\ 
  + Vis + DSDS (Ours)  & 15.01 & 0.5028 & 0.2840 & 0.41 & 0.35 & 59.15 & 52.75 & 0.25 & 0.72 \\ 

        \bottomrule
    \end{tabular}
    \caption{\textbf{``century" capture quantitative evaluation.} Per scene results that compare Nerfacto with baselines (that uses geometric handcrafted regularizers) and Nerfbusters (that uses a data-driven local 3D prior). We highlight that all baselines use the proposed visibility loss. \cref{sec:evaluation_procedure} describes metrics in more detail, $\uparrow$/$\downarrow$ indicates if higher/lower is better.}
    \label{tab:century}
\end{table*}

\begin{table*}[]
    \centering
    \begin{tabular}{l|lll|ll|lllll|l} 
        \toprule
        
 & PSNR $\uparrow$ & SSIM $\uparrow$ & LPIPS $\downarrow$ & Depth $\downarrow$ & Disp. $\downarrow$ & Mean $^{\circ}$ $\downarrow$ & Median $^{\circ}$ $\downarrow$ & \% $30^{\circ}$ $\uparrow$ & Coverage $\uparrow$ \\
        \midrule
        
 Nerfacto Pseudo GT    & 26.67 & 0.8687 & 0.1044 & 0.00 & 0.00 & 0.01 & 0.00 & 1.00 & 0.97 \\ \midrule
 Nerfacto              & 15.68 & 0.4654 & 0.4413 & 14.28 & 0.22 & 65.45 & 60.02 & 0.20 & 0.93 \\ 
  + Visibility Loss    & 15.77 & 0.4691 & 0.4322 & 17.25 & 0.12 & 65.35 & 59.92 & 0.21 & 0.93 \\ 
  + Vis + Sparsity     & 15.67 & 0.4667 & 0.4309 & 17.21 & 0.12 & 65.31 & 59.87 & 0.21 & 0.93 \\ 
  + Vis + TV           & 15.47 & 0.4718 & 0.4314 & 13.03 & 0.11 & 68.00 & 62.88 & 0.19 & 0.91 \\ 
  + Vis + RegNeRF      & 15.67 & 0.4649 & 0.4356 & 13.85 & 0.12 & 65.68 & 60.26 & 0.20 & 0.93 \\ 
  + Vis + DSDS (Ours)  & 15.52 & 0.5086 & 0.3156 & 5.77 & 0.08 & 61.18 & 54.47 & 0.24 & 0.62 \\ 

        \bottomrule
    \end{tabular}
    \caption{\textbf{``flowers" capture quantitative evaluation.} Per scene results that compare Nerfacto with baselines (that uses geometric handcrafted regularizers) and Nerfbusters (that uses a data-driven local 3D prior). We highlight that all baselines use the proposed visibility loss. \cref{sec:evaluation_procedure} describes metrics in more detail, $\uparrow$/$\downarrow$ indicates if higher/lower is better.}
    \label{tab:flowers}
\end{table*}

\begin{table*}[]
    \centering
    \begin{tabular}{l|lll|ll|lllll|l} 
        \toprule
        
 & PSNR $\uparrow$ & SSIM $\uparrow$ & LPIPS $\downarrow$ & Depth $\downarrow$ & Disp. $\downarrow$ & Mean $^{\circ}$ $\downarrow$ & Median $^{\circ}$ $\downarrow$ & \% $30^{\circ}$ $\uparrow$ & Coverage $\uparrow$ \\
        \midrule
        
 Nerfacto Pseudo GT    & 24.09 & 0.8151 & 0.1424 & 0.00 & 0.00 & 0.01 & 0.00 & 1.00 & 1.00 \\ \midrule
 Nerfacto              & 14.86 & 0.4047 & 0.5072 & 0.09 & 10.81 & 63.75 & 58.69 & 0.25 & 1.00 \\ 
  + Visibility Loss    & 14.91 & 0.4197 & 0.4973 & 0.02 & 6.65 & 62.67 & 57.31 & 0.25 & 1.00 \\ 
  + Vis + Sparsity     & 15.04 & 0.4193 & 0.4977 & 0.02 & 9.29 & 62.27 & 56.90 & 0.26 & 1.00 \\ 
  + Vis + TV           & 15.20 & 0.4212 & 0.5110 & 0.04 & 1.18 & 73.74 & 70.17 & 0.15 & 1.00 \\ 
  + Vis + RegNeRF      & 15.13 & 0.4205 & 0.4926 & 0.03 & 5.42 & 63.18 & 57.80 & 0.25 & 1.00 \\ 
  + Vis + DSDS (Ours)  & 15.86 & 0.4466 & 0.3726 & 0.00 & 0.11 & 53.09 & 45.57 & 0.31 & 0.63 \\ 

        \bottomrule
    \end{tabular}
    \caption{\textbf{``garbage" capture quantitative evaluation.} Per scene results that compare Nerfacto with baselines (that uses geometric handcrafted regularizers) and Nerfbusters (that uses a data-driven local 3D prior). We highlight that all baselines use the proposed visibility loss. \cref{sec:evaluation_procedure} describes metrics in more detail, $\uparrow$/$\downarrow$ indicates if higher/lower is better.}
    \label{tab:garbage}
\end{table*}

\begin{table*}[]
    \centering
    \begin{tabular}{l|lll|ll|lllll|l} 
        \toprule
        
 & PSNR $\uparrow$ & SSIM $\uparrow$ & LPIPS $\downarrow$ & Depth $\downarrow$ & Disp. $\downarrow$ & Mean $^{\circ}$ $\downarrow$ & Median $^{\circ}$ $\downarrow$ & \% $30^{\circ}$ $\uparrow$ & Coverage $\uparrow$ \\
        \midrule
        
 Nerfacto Pseudo GT    & 23.44 & 0.7616 & 0.1774 & 0.00 & 0.00 & 0.01 & 0.00 & 1.00 & 0.95 \\ \midrule
 Nerfacto              & 15.99 & 0.3123 & 0.5019 & 0.12 & 1.17 & 61.20 & 55.39 & 0.24 & 0.96 \\ 
  + Visibility Loss    & 15.93 & 0.3117 & 0.4992 & 0.11 & 0.90 & 60.57 & 54.66 & 0.25 & 0.96 \\ 
  + Vis + Sparsity     & 15.97 & 0.3145 & 0.4992 & 0.10 & 0.77 & 60.91 & 55.02 & 0.25 & 0.96 \\ 
  + Vis + TV           & 16.03 & 0.3168 & 0.5123 & 0.11 & 0.76 & 68.94 & 63.99 & 0.19 & 0.96 \\ 
  + Vis + RegNeRF      & 15.77 & 0.3014 & 0.5149 & 0.14 & 1.12 & 62.16 & 56.47 & 0.23 & 0.96 \\ 
  + Vis + DSDS (Ours)  & 15.72 & 0.4560 & 0.3074 & 0.11 & 0.12 & 54.32 & 46.71 & 0.31 & 0.58 \\ 

        \bottomrule
    \end{tabular}
    \caption{\textbf{``picnic" capture quantitative evaluation.} Per scene results that compare Nerfacto with baselines (that uses geometric handcrafted regularizers) and Nerfbusters (that uses a data-driven local 3D prior). We highlight that all baselines use the proposed visibility loss. \cref{sec:evaluation_procedure} describes metrics in more detail, $\uparrow$/$\downarrow$ indicates if higher/lower is better.}
    \label{tab:picnic}
\end{table*}

\begin{table*}[]
    \centering
    \begin{tabular}{l|lll|ll|lllll|l} 
        \toprule
        
 & PSNR $\uparrow$ & SSIM $\uparrow$ & LPIPS $\downarrow$ & Depth $\downarrow$ & Disp. $\downarrow$ & Mean $^{\circ}$ $\downarrow$ & Median $^{\circ}$ $\downarrow$ & \% $30^{\circ}$ $\uparrow$ & Coverage $\uparrow$ \\
        \midrule
        
 Nerfacto Pseudo GT    & 31.69 & 0.9606 & 0.0399 & 0.00 & 0.00 & 0.01 & 0.00 & 1.00 & 0.92 \\ \midrule
 Nerfacto              & 20.31 & 0.6903 & 0.3267 & 0.60 & 0.83 & 56.86 & 48.00 & 0.29 & 0.93 \\ 
  + Visibility Loss    & 25.83 & 0.8837 & 0.0874 & 0.01 & 0.01 & 42.55 & 34.20 & 0.44 & 0.77 \\ 
  + Vis + Sparsity     & 25.79 & 0.8806 & 0.0909 & 0.00 & 0.01 & 42.59 & 34.28 & 0.44 & 0.77 \\ 
  + Vis + TV           & 26.23 & 0.9000 & 0.0867 & 0.00 & 0.01 & 47.29 & 40.29 & 0.36 & 0.71 \\ 
  + Vis + RegNeRF      & 25.24 & 0.8734 & 0.0970 & 0.01 & 0.01 & 42.62 & 34.26 & 0.44 & 0.77 \\ 
  + Vis + DSDS (Ours)  & 25.71 & 0.9048 & 0.0502 & 0.00 & 0.00 & 45.90 & 39.17 & 0.36 & 0.28 \\ 

        \bottomrule
    \end{tabular}
    \caption{\textbf{``pikachu" capture quantitative evaluation.} Per scene results that compare Nerfacto with baselines (that uses geometric handcrafted regularizers) and Nerfbusters (that uses a data-driven local 3D prior). We highlight that all baselines use the proposed visibility loss. \cref{sec:evaluation_procedure} describes metrics in more detail, $\uparrow$/$\downarrow$ indicates if higher/lower is better.}
    \label{tab:pikachu}
\end{table*}

\begin{table*}[]
    \centering
    \begin{tabular}{l|lll|ll|lllll|l} 
        \toprule
        
 & PSNR $\uparrow$ & SSIM $\uparrow$ & LPIPS $\downarrow$ & Depth $\downarrow$ & Disp. $\downarrow$ & Mean $^{\circ}$ $\downarrow$ & Median $^{\circ}$ $\downarrow$ & \% $30^{\circ}$ $\uparrow$ & Coverage $\uparrow$ \\
        \midrule
        
 Nerfacto Pseudo GT    & 23.38 & 0.8008 & 0.1384 & 0.00 & 0.00 & 0.01 & 0.00 & 1.00 & 0.93 \\ \midrule
 Nerfacto              & 19.62 & 0.5910 & 0.2673 & 0.08 & 0.07 & 59.62 & 53.54 & 0.25 & 0.93 \\ 
  + Visibility Loss    & 19.61 & 0.5914 & 0.2637 & 0.08 & 0.07 & 59.27 & 53.12 & 0.25 & 0.93 \\ 
  + Vis + Sparsity     & 19.67 & 0.5906 & 0.2697 & 0.07 & 0.07 & 60.28 & 54.29 & 0.24 & 0.93 \\ 
  + Vis + TV           & 19.64 & 0.5891 & 0.2793 & 0.07 & 0.07 & 65.09 & 59.67 & 0.20 & 0.93 \\ 
  + Vis + RegNeRF      & 19.62 & 0.5917 & 0.2665 & 0.15 & 0.07 & 59.46 & 53.35 & 0.25 & 0.93 \\ 
  + Vis + DSDS (Ours)  & 19.23 & 0.6165 & 0.2387 & 0.08 & 0.08 & 58.58 & 52.35 & 0.26 & 0.82 \\ 

        \bottomrule
    \end{tabular}
    \caption{\textbf{``pipe" capture quantitative evaluation.} Per scene results that compare Nerfacto with baselines (that uses geometric handcrafted regularizers) and Nerfbusters (that uses a data-driven local 3D prior). We highlight that all baselines use the proposed visibility loss. \cref{sec:evaluation_procedure} describes metrics in more detail, $\uparrow$/$\downarrow$ indicates if higher/lower is better.}
    \label{tab:pipe}
\end{table*}

\begin{table*}[]
    \centering
    \begin{tabular}{l|lll|ll|lllll|l} 
        \toprule
        
 & PSNR $\uparrow$ & SSIM $\uparrow$ & LPIPS $\downarrow$ & Depth $\downarrow$ & Disp. $\downarrow$ & Mean $^{\circ}$ $\downarrow$ & Median $^{\circ}$ $\downarrow$ & \% $30^{\circ}$ $\uparrow$ & Coverage $\uparrow$ \\
        \midrule
        
 Nerfacto Pseudo GT    & 29.66 & 0.9349 & 0.0476 & 0.00 & 0.00 & 0.01 & 0.00 & 1.00 & 0.57 \\ \midrule
 Nerfacto              & 16.42 & 0.6146 & 0.3479 & 229.88 & 0.38 & 74.54 & 70.76 & 0.17 & 0.71 \\ 
  + Visibility Loss    & 20.55 & 0.7090 & 0.1824 & 40.48 & 0.05 & 60.30 & 52.21 & 0.28 & 0.39 \\ 
  + Vis + Sparsity     & 20.48 & 0.7066 & 0.1867 & 40.85 & 0.05 & 59.98 & 51.79 & 0.28 & 0.39 \\ 
  + Vis + TV           & 20.61 & 0.7148 & 0.1723 & 36.89 & 0.05 & 60.68 & 52.63 & 0.27 & 0.39 \\ 
  + Vis + RegNeRF      & 20.37 & 0.7038 & 0.1866 & 40.52 & 0.05 & 60.35 & 52.25 & 0.28 & 0.39 \\ 
  + Vis + DSDS (Ours)  & 20.20 & 0.7535 & 0.1254 & 232.09 & 0.03 & 55.40 & 46.52 & 0.32 & 0.25 \\ 

        \bottomrule
    \end{tabular}
    \caption{\textbf{``plant" capture quantitative evaluation.} Per scene results that compare Nerfacto with baselines (that uses geometric handcrafted regularizers) and Nerfbusters (that uses a data-driven local 3D prior). We highlight that all baselines use the proposed visibility loss. \cref{sec:evaluation_procedure} describes metrics in more detail, $\uparrow$/$\downarrow$ indicates if higher/lower is better.}
    \label{tab:plant}
\end{table*}

\begin{table*}[]
    \centering
    \begin{tabular}{l|lll|ll|lllll|l} 
        \toprule
        
 & PSNR $\uparrow$ & SSIM $\uparrow$ & LPIPS $\downarrow$ & Depth $\downarrow$ & Disp. $\downarrow$ & Mean $^{\circ}$ $\downarrow$ & Median $^{\circ}$ $\downarrow$ & \% $30^{\circ}$ $\uparrow$ & Coverage $\uparrow$ \\
        \midrule
        
 Nerfacto Pseudo GT    & 29.51 & 0.9186 & 0.0554 & 0.00 & 0.00 & 0.01 & 0.00 & 1.00 & 0.91 \\ \midrule
 Nerfacto              & 19.96 & 0.7159 & 0.2383 & 1.23 & 0.02 & 47.34 & 40.24 & 0.36 & 0.92 \\ 
  + Visibility Loss    & 19.94 & 0.7173 & 0.2412 & 0.64 & 0.02 & 46.67 & 39.59 & 0.37 & 0.92 \\ 
  + Vis + Sparsity     & 19.92 & 0.7195 & 0.2416 & 0.64 & 0.02 & 46.34 & 39.34 & 0.37 & 0.92 \\ 
  + Vis + TV           & 19.94 & 0.7190 & 0.2428 & 0.69 & 0.02 & 47.26 & 40.16 & 0.36 & 0.92 \\ 
  + Vis + RegNeRF      & 19.83 & 0.7168 & 0.2424 & 0.61 & 0.02 & 47.21 & 40.14 & 0.36 & 0.92 \\ 
  + Vis + DSDS (Ours)  & 19.14 & 0.7149 & 0.2076 & 0.37 & 0.03 & 47.27 & 40.45 & 0.35 & 0.87 \\ 

        \bottomrule
    \end{tabular}
    \caption{\textbf{``roses" capture quantitative evaluation.} Per scene results that compare Nerfacto with baselines (that uses geometric handcrafted regularizers) and Nerfbusters (that uses a data-driven local 3D prior). We highlight that all baselines use the proposed visibility loss. \cref{sec:evaluation_procedure} describes metrics in more detail, $\uparrow$/$\downarrow$ indicates if higher/lower is better.}
    \label{tab:roses}
\end{table*}

\begin{table*}[]
    \centering
    \begin{tabular}{l|lll|ll|lllll|l} 
        \toprule
        
 & PSNR $\uparrow$ & SSIM $\uparrow$ & LPIPS $\downarrow$ & Depth $\downarrow$ & Disp. $\downarrow$ & Mean $^{\circ}$ $\downarrow$ & Median $^{\circ}$ $\downarrow$ & \% $30^{\circ}$ $\uparrow$ & Coverage $\uparrow$ \\
        \midrule
        
 Nerfacto Pseudo GT    & 27.30 & 0.9180 & 0.0784 & 0.00 & 0.00 & 0.01 & 0.00 & 1.00 & 0.99 \\ \midrule
 Nerfacto              & 17.60 & 0.6877 & 0.3259 & 0.04 & 0.47 & 52.08 & 43.98 & 0.33 & 0.89 \\ 
  + Visibility Loss    & 17.51 & 0.6878 & 0.3294 & 0.04 & 0.48 & 51.82 & 43.62 & 0.34 & 0.88 \\ 
  + Vis + Sparsity     & 17.35 & 0.6818 & 0.3330 & 0.04 & 0.48 & 51.86 & 43.71 & 0.33 & 0.88 \\ 
  + Vis + TV           & 16.49 & 0.6895 & 0.3311 & 0.04 & 0.41 & 54.34 & 46.63 & 0.31 & 0.86 \\ 
  + Vis + RegNeRF      & 17.39 & 0.6832 & 0.3318 & 0.04 & 0.48 & 51.89 & 43.67 & 0.34 & 0.88 \\ 
  + Vis + DSDS (Ours)  & 18.23 & 0.7060 & 0.2867 & 0.02 & 0.28 & 48.67 & 41.31 & 0.35 & 0.69 \\ 

        \bottomrule
    \end{tabular}
    \caption{\textbf{``table" capture quantitative evaluation.} Per scene results that compare Nerfacto with baselines (that uses geometric handcrafted regularizers) and Nerfbusters (that uses a data-driven local 3D prior). We highlight that all baselines use the proposed visibility loss. \cref{sec:evaluation_procedure} describes metrics in more detail, $\uparrow$/$\downarrow$ indicates if higher/lower is better.}
    \label{tab:table}
\end{table*}


{\small
\bibliographystyle{ieee_fullname}
\bibliography{egbib}

\begin{thebibliography}{10}\itemsep=-1pt

\bibitem{barron2021mip}
Jonathan~T Barron, Ben Mildenhall, Matthew Tancik, Peter Hedman, Ricardo
  Martin-Brualla, and Pratul~P Srinivasan.
\newblock Mip-nerf: A multiscale representation for anti-aliasing neural
  radiance fields.
\newblock In {\em Proceedings of the IEEE/CVF International Conference on
  Computer Vision}, pages 5855--5864, 2021.

\bibitem{barron2022mip}
Jonathan~T Barron, Ben Mildenhall, Dor Verbin, Pratul~P Srinivasan, and Peter
  Hedman.
\newblock Mip-nerf 360: Unbounded anti-aliased neural radiance fields.
\newblock In {\em Proceedings of the IEEE/CVF Conference on Computer Vision and
  Pattern Recognition}, pages 5470--5479, 2022.

\bibitem{chabra2020deep}
Rohan Chabra, Jan~E Lenssen, Eddy Ilg, Tanner Schmidt, Julian Straub, Steven
  Lovegrove, and Richard Newcombe.
\newblock Deep local shapes: Learning local sdf priors for detailed 3d
  reconstruction.
\newblock In {\em Computer Vision--ECCV 2020: 16th European Conference,
  Glasgow, UK, August 23--28, 2020, Proceedings, Part XXIX 16}, pages 608--625.
  Springer, 2020.

\bibitem{chang2015shapenet}
Angel~X Chang, Thomas Funkhouser, Leonidas Guibas, Pat Hanrahan, Qixing Huang,
  Zimo Li, Silvio Savarese, Manolis Savva, Shuran Song, Hao Su, et~al.
\newblock Shapenet: An information-rich 3d model repository.
\newblock {\em arXiv preprint arXiv:1512.03012}, 2015.

\bibitem{chen2022tensorf}
Anpei Chen, Zexiang Xu, Andreas Geiger, Jingyi Yu, and Hao Su.
\newblock Tensorf: Tensorial radiance fields.
\newblock In {\em Computer Vision--ECCV 2022: 17th European Conference, Tel
  Aviv, Israel, October 23--27, 2022, Proceedings, Part XXXII}, pages 333--350.
  Springer, 2022.

\bibitem{fridovich2023k}
Sara Fridovich-Keil, Giacomo Meanti, Frederik Warburg, Benjamin Recht, and
  Angjoo Kanazawa.
\newblock K-planes: Explicit radiance fields in space, time, and appearance.
\newblock {\em arXiv preprint arXiv:2301.10241}, 2023.

\bibitem{fridovich2022plenoxels}
Sara Fridovich-Keil, Alex Yu, Matthew Tancik, Qinhong Chen, Benjamin Recht, and
  Angjoo Kanazawa.
\newblock Plenoxels: Radiance fields without neural networks.
\newblock In {\em Proceedings of the IEEE/CVF Conference on Computer Vision and
  Pattern Recognition}, pages 5501--5510, 2022.

\bibitem{gao2022monocular}
Hang Gao, Ruilong Li, Shubham Tulsiani, Bryan Russell, and Angjoo Kanazawa.
\newblock Monocular dynamic view synthesis: A reality check.
\newblock In {\em Advances in Neural Information Processing Systems}, 2022.

\bibitem{ho2020denoising}
Jonathan Ho, Ajay Jain, and Pieter Abbeel.
\newblock Denoising diffusion probabilistic models.
\newblock {\em Advances in Neural Information Processing Systems},
  33:6840--6851, 2020.

\bibitem{jensen2014large}
Rasmus Jensen, Anders Dahl, George Vogiatzis, Engil Tola, and Henrik Aan{\ae}s.
\newblock Large scale multi-view stereopsis evaluation.
\newblock In {\em 2014 IEEE Conference on Computer Vision and Pattern
  Recognition}, pages 406--413. IEEE, 2014.

\bibitem{jin2021image}
Yuhe Jin, Dmytro Mishkin, Anastasiia Mishchuk, Jiri Matas, Pascal Fua,
  Kwang~Moo Yi, and Eduard Trulls.
\newblock Image matching across wide baselines: From paper to practice.
\newblock {\em International Journal of Computer Vision}, 129(2):517--547,
  2021.

\bibitem{li2022infinitenature}
Zhengqi Li, Qianqian Wang, Noah Snavely, and Angjoo Kanazawa.
\newblock Infinitenature-zero: Learning perpetual view generation of natural
  scenes from single images.
\newblock In {\em Computer Vision--ECCV 2022: 17th European Conference, Tel
  Aviv, Israel, October 23--27, 2022, Proceedings, Part I}, pages 515--534.
  Springer, 2022.

\bibitem{lin2021barf}
Chen-Hsuan Lin, Wei-Chiu Ma, Antonio Torralba, and Simon Lucey.
\newblock Barf: Bundle-adjusting neural radiance fields.
\newblock In {\em Proceedings of the IEEE/CVF International Conference on
  Computer Vision}, pages 5741--5751, 2021.

\bibitem{liu2021infinite}
Andrew Liu, Richard Tucker, Varun Jampani, Ameesh Makadia, Noah Snavely, and
  Angjoo Kanazawa.
\newblock Infinite nature: Perpetual view generation of natural scenes from a
  single image.
\newblock In {\em Proceedings of the IEEE/CVF International Conference on
  Computer Vision}, pages 14458--14467, 2021.

\bibitem{martin2021nerf}
Ricardo Martin-Brualla, Noha Radwan, Mehdi~SM Sajjadi, Jonathan~T Barron,
  Alexey Dosovitskiy, and Daniel Duckworth.
\newblock Nerf in the wild: Neural radiance fields for unconstrained photo
  collections.
\newblock In {\em Proceedings of the IEEE/CVF Conference on Computer Vision and
  Pattern Recognition}, pages 7210--7219, 2021.

\bibitem{max1995}
N. Max.
\newblock Optical models for direct volume rendering.
\newblock {\em IEEE Transactions on Visualization and Computer Graphics},
  1(2):99--108, 1995.

\bibitem{melas2023realfusion}
Luke Melas-Kyriazi, Christian Rupprecht, Iro Laina, and Andrea Vedaldi.
\newblock Realfusion: 360 $\{$$\backslash$deg$\}$ reconstruction of any object
  from a single image.
\newblock {\em arXiv preprint arXiv:2302.10663}, 2023.

\bibitem{mildenhall2019local}
Ben Mildenhall, Pratul~P Srinivasan, Rodrigo Ortiz-Cayon, Nima~Khademi
  Kalantari, Ravi Ramamoorthi, Ren Ng, and Abhishek Kar.
\newblock Local light field fusion: Practical view synthesis with prescriptive
  sampling guidelines.
\newblock {\em ACM Transactions on Graphics (TOG)}, 38(4):1--14, 2019.

\bibitem{mildenhall2019llff}
Ben Mildenhall, Pratul~P. Srinivasan, Rodrigo Ortiz-Cayon, Nima~Khademi
  Kalantari, Ravi Ramamoorthi, Ren Ng, and Abhishek Kar.
\newblock Local light field fusion: Practical view synthesis with prescriptive
  sampling guidelines.
\newblock {\em ACM Transactions on Graphics (TOG)}, 2019.

\bibitem{mildenhall2021nerf}
Ben Mildenhall, Pratul~P Srinivasan, Matthew Tancik, Jonathan~T Barron, Ravi
  Ramamoorthi, and Ren Ng.
\newblock Nerf: Representing scenes as neural radiance fields for view
  synthesis.
\newblock {\em Communications of the ACM}, 65(1):99--106, 2021.

\bibitem{mittal2022autosdf}
Paritosh Mittal, Yen-Chi Cheng, Maneesh Singh, and Shubham Tulsiani.
\newblock Autosdf: Shape priors for 3d completion, reconstruction and
  generation.
\newblock In {\em Proceedings of the IEEE/CVF Conference on Computer Vision and
  Pattern Recognition}, pages 306--315, 2022.

\bibitem{muller2022diffrf}
Norman M{\"u}ller, Yawar Siddiqui, Lorenzo Porzi, Samuel~Rota Bul{\`o}, Peter
  Kontschieder, and Matthias Nie{\ss}ner.
\newblock Diffrf: Rendering-guided 3d radiance field diffusion.
\newblock {\em arXiv preprint arXiv:2212.01206}, 2022.

\bibitem{muller2022instant}
Thomas M{\"u}ller, Alex Evans, Christoph Schied, and Alexander Keller.
\newblock Instant neural graphics primitives with a multiresolution hash
  encoding.
\newblock {\em ACM Transactions on Graphics (ToG)}, 41(4):1--15, 2022.

\bibitem{nichol2021improved}
Alexander~Quinn Nichol and Prafulla Dhariwal.
\newblock Improved denoising diffusion probabilistic models.
\newblock In {\em International Conference on Machine Learning}, pages
  8162--8171. PMLR, 2021.

\bibitem{Niemeyer_2022_CVPR}
Michael Niemeyer, Jonathan~T. Barron, Ben Mildenhall, Mehdi S.~M. Sajjadi,
  Andreas Geiger, and Noha Radwan.
\newblock Regnerf: Regularizing neural radiance fields for view synthesis from
  sparse inputs.
\newblock In {\em Proceedings of the IEEE/CVF Conference on Computer Vision and
  Pattern Recognition (CVPR)}, pages 5480--5490, June 2022.

\bibitem{park2019deepsdf}
Jeong~Joon Park, Peter Florence, Julian Straub, Richard Newcombe, and Steven
  Lovegrove.
\newblock Deepsdf: Learning continuous signed distance functions for shape
  representation.
\newblock In {\em Proceedings of the IEEE/CVF conference on computer vision and
  pattern recognition}, pages 165--174, 2019.

\bibitem{poole2022dreamfusion}
Ben Poole, Ajay Jain, Jonathan~T Barron, and Ben Mildenhall.
\newblock Dreamfusion: Text-to-3d using 2d diffusion.
\newblock {\em arXiv preprint arXiv:2209.14988}, 2022.

\bibitem{razavi2019generating}
Ali Razavi, Aaron Van~den Oord, and Oriol Vinyals.
\newblock Generating diverse high-fidelity images with vq-vae-2.
\newblock {\em Advances in neural information processing systems}, 32, 2019.

\bibitem{reizenstein21co3d}
Jeremy Reizenstein, Roman Shapovalov, Philipp Henzler, Luca Sbordone, Patrick
  Labatut, and David Novotny.
\newblock Common objects in 3d: Large-scale learning and evaluation of
  real-life 3d category reconstruction.
\newblock In {\em International Conference on Computer Vision}, 2021.

\bibitem{rombach2021highresolution}
Robin Rombach, Andreas Blattmann, Dominik Lorenz, Patrick Esser, and Björn
  Ommer.
\newblock High-resolution image synthesis with latent diffusion models, 2021.

\bibitem{ronneberger2015u}
Olaf Ronneberger, Philipp Fischer, and Thomas Brox.
\newblock U-net: Convolutional networks for biomedical image segmentation.
\newblock In {\em Medical Image Computing and Computer-Assisted
  Intervention--MICCAI 2015: 18th International Conference, Munich, Germany,
  October 5-9, 2015, Proceedings, Part III 18}, pages 234--241. Springer, 2015.

\bibitem{sabour2023robustnerf}
Sara Sabour, Suhani Vora, Daniel Duckworth, Ivan Krasin, David~J Fleet, and
  Andrea Tagliasacchi.
\newblock Robustnerf: Ignoring distractors with robust losses.
\newblock {\em arXiv preprint arXiv:2302.00833}, 2023.

\bibitem{sohl2015deep}
Jascha Sohl-Dickstein, Eric Weiss, Niru Maheswaranathan, and Surya Ganguli.
\newblock Deep unsupervised learning using nonequilibrium thermodynamics.
\newblock In {\em International Conference on Machine Learning}, pages
  2256--2265. PMLR, 2015.

\bibitem{song2019generative}
Yang Song and Stefano Ermon.
\newblock Generative modeling by estimating gradients of the data distribution.
\newblock {\em Advances in neural information processing systems}, 32, 2019.

\bibitem{tancik2023nerfstudio}
Matthew Tancik, Ethan Weber, Evonne Ng, Ruilong Li, Brent Yi, Justin Kerr,
  Terrance Wang, Alexander Kristoffersen, Jake Austin, Kamyar Salahi, et~al.
\newblock Nerfstudio: A modular framework for neural radiance field
  development.
\newblock {\em arXiv preprint arXiv:2302.04264}, 2023.

\bibitem{van2017neural}
Aaron Van Den~Oord, Oriol Vinyals, et~al.
\newblock Neural discrete representation learning.
\newblock {\em Advances in neural information processing systems}, 30, 2017.

\bibitem{wang2022score}
Haochen Wang, Xiaodan Du, Jiahao Li, Raymond~A Yeh, and Greg Shakhnarovich.
\newblock Score jacobian chaining: Lifting pretrained 2d diffusion models for
  3d generation.
\newblock {\em arXiv preprint arXiv:2212.00774}, 2022.

\bibitem{wang2021nerf}
Zirui Wang, Shangzhe Wu, Weidi Xie, Min Chen, and Victor~Adrian Prisacariu.
\newblock Nerf--: Neural radiance fields without known camera parameters.
\newblock {\em arXiv preprint arXiv:2102.07064}, 2021.

\bibitem{warburg2022self}
Frederik Warburg, Daniel Hernandez-Juarez, Juan Tarrio, Alexander Vakhitov,
  Ujwal Bonde, and Pablo~F Alcantarilla.
\newblock Self-supervised depth completion for active stereo.
\newblock {\em IEEE Robotics and Automation Letters}, 7(2):3475--3482, 2022.

\bibitem{warburg2022sparseformer}
Frederik Warburg, Michael Ramamonjisoa, and Manuel L{\'o}pez-Antequera.
\newblock Sparseformer: Attention-based depth completion network.
\newblock {\em arXiv preprint arXiv:2206.04557}, 2022.

\bibitem{wynn-2023-diffusionerf}
Jamie Wynn and Daniyar Turmukhambetov.
\newblock Diffusionerf: Regularizing neural radiance fields with denoising
  diffusion models.
\newblock In {\em arxiv}, 2023.

\bibitem{yu2021plenoctrees}
Alex Yu, Ruilong Li, Matthew Tancik, Hao Li, Ren Ng, and Angjoo Kanazawa.
\newblock Plenoctrees for real-time rendering of neural radiance fields.
\newblock In {\em Proceedings of the IEEE/CVF International Conference on
  Computer Vision}, pages 5752--5761, 2021.

\bibitem{yu2021pixelnerf}
Alex Yu, Vickie Ye, Matthew Tancik, and Angjoo Kanazawa.
\newblock pixelnerf: Neural radiance fields from one or few images.
\newblock In {\em Proceedings of the IEEE/CVF Conference on Computer Vision and
  Pattern Recognition}, pages 4578--4587, 2021.

\bibitem{zhang2018deep}
Yinda Zhang and Thomas Funkhouser.
\newblock Deep depth completion of a single rgb-d image.
\newblock In {\em Proceedings of the IEEE Conference on Computer Vision and
  Pattern Recognition}, pages 175--185, 2018.

\bibitem{zhou2022sparsefusion}
Zhizhuo Zhou and Shubham Tulsiani.
\newblock Sparsefusion: Distilling view-conditioned diffusion for 3d
  reconstruction.
\newblock {\em arXiv preprint arXiv:2212.00792}, 2022.

\end{thebibliography}
}

\end{document}